# MODEL-BASED IDENTIFICATION AND CONTROL OF A ONE-LEGGED HOPPING ROBOT

A THESIS SUBMITTED TO

THE GRADUATE SCHOOL OF ENGINEERING AND SCIENCE

OF BILKENT UNIVERSITY

IN PARTIAL FULFILLMENT OF THE REQUIREMENTS FOR

THE DEGREE OF

MASTER OF SCIENCE

IN

ELECTRICAL AND ELECTRONICS ENGINEERING

By
Hasan Eftun Orhon
January 2018

Model-Based Identification and Control of a One-Legged Hopping Robot
By Hasan Eftun Orhon
January 2018

We certify that we have read this thesis and that in our opinion it is fully adequate, in scope and in quality, as a thesis for the degree of Master of Science.

---
Ömer Morgül(Advisor)

---
Uluç Saranlı

---
Melih Çakmakcı

Approved for the Graduate School of Engineering and Science:

---
Ezhan Karaşan
Director of the Graduate School

# ABSTRACT

# MODEL-BASED IDENTIFICATION AND CONTROL OF A ONE-LEGGED HOPPING ROBOT

Hasan Eftun Orhon
M.S. in Electrical and Electronics Engineering
Advisor: Ömer Morgül
January 2018

Spring-mass models are well established tools for the analysis and control of legged locomotion. Among the alternatives, spring-loaded inverted pendulum (SLIP) model has shown to be a very accurate descriptor of animal locomotion. Despite its wide use, the SLIP model includes non-integrable stance dynamics that prevent analytical solutions for its equations of motion. Fortunately, there are approximate analytical solutions for different SLIP variants. However, the practicality of such approximations are mostly tested on simulation studies with a few notable exceptions.

This thesis extends upon a recent approximation to a hip torque actuated dissipative SLIP (TD-SLIP) model that uses torque actuation to compensate for energy losses. Systematic experiments for careful assessment of the predictive performance of the approximate analytical solution is presented on a well-instrumented one-legged hopping robot which is revised to enhance compatibility and accuracy of the system. Electronic structure of the robot is modified according to TD-SLIP model such that robot uses a real-time operating system to increase processing speed. Using the parameters and results generated by the predictive performance of the approximate analytical solution, a model-based controller is designed and implemented on the robot platform to generate a stable closed-loop running behaviour on the one legged hoping robot platform. In addition, ground reaction forces during the stance phase on the experimental platform is investigated and compared with the human running and the traditional SLIP model data to understand if torque-actuated models approximate natural locomotion better than traditional model.

# ÖZET

# TEK-BACAKLI ZIPLAYAN ROBOT ÜZERINDE MODEL TABANLI TANIMLAMA VE KONTROL

Hasan Eftun Orhon
Elektrik Elektronik Mühendisliği, Yüksek Lisans
Tez Danışmanı: Ömer Morgül
Ocak 2018

Yay-kütle modelleri bacaklı hareket sistemlerini incelemek ve kontrol etmek için sıkça kullanılan bir gereçtir. Alternatifleri arasında, yaylı ters sarkaç (YTS) modelinin canlı hareketlerini oldukça doğru bir şekilde açıkladığı görülmektedir. Geniş kullanım alanına rağmen, YTS modeli integrali alınamayan, bu nedenle analitik olarak çözülemeyen, hareket denklemlerine sahiptir. Neyse ki, birçok farklı YTS modeli için geliştirilmiş yakınsamalı analitik çözümler literatürde mevcuttur. Ancak bu yakınsamaların kullanışlılığı, birkaç örnek dışında, genelde benzetim ortamlarında test edilmektedir.

Bu tez çalışmasında yakın bir zamanda geliştirilen, sistemde gerçekleşen enerji kaybını kalça torku ile telafi eden, torklu sönümlemeli yaylı ters sarkaç modelinin (TS-YTS) kapsamı genişletilmiştir. TS-YTS modelinin yakınsamalı analitik çözümünün kestirimci performans analizi bu çalışma için geliştirilen tek bacaklı zıplayan robot üzerinde sistematik deneylerle değerlendirilmiştir. Bu robotun elektronik alt-yapısı TS-YTS modeline uygun olacak şekilde işlem hızını artırmak için gerçek zamanlı bir işletim sistemi üzerine kurulmuştur. Kestirimci performans analizinin sonuçlarını ve burdan çıkan sistem parametrelerini kullanarak, robot üzerinde kararlı koşu davranışını gözlemleyebilmek için model-tabanlı bir kontrolcü tasarlanmış ve uygulanmıştır. Bunlara ek olarak, geliştirilen robotun yere basma fazı boyunca gözlemlenen yer tepki kuvveti incelenerek TS-TYS modeli, insan koşma hareketi ve geleneksel YTS modelleri ile karşılaştırılmış ve tork kullanan YTS modellerinin bu doğal hareketi daha iyi tahmin edip edemediği test edilmiştir.

*Anahtar sözcükler*: Bacaklı hareket, YTS modeli, Model-tabanlı kontrolcü, Yer tepki kuvveti, Doğadan esinlenmiş robotlar, Yakınsalamı analitik çözüm, Gerçek-zamanlı işletim sistemleri.

# Acknowledgement

My journey of master for three years has been an astonishing experience that cannot be possible without the inspiration and the support of many people.

Firstly, I would like to express my deepest gratitude for my supervisor, Ömer Morgül for his unlimited guidance and encouragement about my work. I would like to offer my special thanks to Uluç Saranlı. I hugely indebted to them for their complete support for my personal and academic works which inspire me on my first steps at becoming an researcher. I will always show my endless respect to them. My grateful thanks also extends to Melih Çakmakçı for approving my work.

There is one more person that guide me through this journey and the others, İsmail Uyanık. With his help and support, I was able to succeed at many things that I do not think whether I will or not.

I am also grateful for Hasan Hamzaçebi for endless nights with the robot that we struggle together.

Additionally, I would like to thank to the member of our research group that are contributed to my personal and academic life. Mustafa Gül, Mansur Arısoy, Dilan Öztürk, Elvan Kuzucu, Bengisu Özbay, Mustafa Oğuz Yeğin, Ali Nail İnal, Caner Odabaş, Bahadır Çatalbaş, Görkem Seçer, Deniz Kerimoğlu, Ahmet Safa Öztürk for the best three years of my life in Bilkent.

Other than my research group there are some friends that both motivate and help me either directly or indirectly. I cannot be more happy to have Özgün Yavuz, Furkan Kökdoğan, Osman Erdem, Deniz Doğan, Sinan Saraçoğlu, Burak Demirel, Faruk Uyar as friends and thanks for all the wonderful time we spend together.

I would like to thank Mürüvet Parlakay for help on administrative work and thanks for their helps on technical issues Ergün Hırlakoğlu, Onur Bostancı and Ufuk Tufan.

I appreciate the financial support provided from the Scientific and Technological

Research Council of Turkey (TÜBİTAK). The work presented in this thesis was supported by TÜBİTAK through projects 114E277, 215E050.

Finally, I would like to thank my family Mahmut Orhon, Nihal Orhon, Ali Orhon, Melis Şakire Tokuç, Özlem Orhon, Ogün Tokuç and all of my nephews and niece for their unconditional love and support for me.

# Contents

# List of Figures

# List of Tables

# Chapter 1

# Introduction

It is a long discussed fact that legged robots perform better on rough terrains due to their ability to choose optimum foothold placement during their locomotion [1]. Motivated by this idea, various modelling, identification and control tools have been developed to analyse and control legged locomotor systems [2–6]. Especially during the last decade, many successful examples are proposed to demonstrate the ability of legged robot platforms on rough terrain locomotion [3,7–13]. These platforms present promising results for the future of legged locomotion.

The main motivation of this thesis work is to develop a model-based controller on the one legged hoping robot in our laboratory see [14]. For this purpose, we examine the torque-actuated spring-loaded inverted pendulum (TD-SLIP) model given in [15] which provides promising result in simulation environment. In this thesis, we focus on experimental validation of the approximate analytical solution of extended TD-SLIP model which will provide a novel basis for implementation of the model-based controller on our revised one legged hoping robot according to TD-SLIP model. In the following sections, we will investigate the existing studies and propose the methodology of this thesis that will provide valuable information to reach our motivations.

## 1.1 Model-Based and Data-Driven Methods for Analyzing Legged Locomotion

There are two main directions for the analysis of legged locomotion; mechanics-based mathematical models and data-driven system identification methods. Data-driven methods aim to obtain input–output models of legged locomotion [16–21]. These methods provide an easy translation between different models by eliminating complex and highly non-linear nature of the legged locomotion systems. Using same control inputs for each stride legged locomotor systems can reach a stable periodic orbit called limit cycle. These methods generally investigate locomotion dynamics around a stable limit cycle. The hybrid nature of legged locomotion can be approximated as a linear time-periodic (LTP) system around its limit cycle [22]. Hence, LTP analyis, identification and control methods in the literature [23–25] can be utilized for legged locomotion models as well. Such system identification methods can provide schemes for a certain set of legged locomotion models as in [26].

On the other hand, there are some mechanics-based mathematical models that considers the principles of dynamics to design feedforward predictors for the analysis and control of legged locomotion [27–30]. In this thesis, a model-based identification and control will be established. Model-based systems directly use mechanical properties and system dynamics of the locomotion model. Despite complexity and non-linearity of hybrid dynamics of the legged locomotion systems, Model-based identifications can offer accurate solutions for this type of systems using simple approximations on the system dynamics as given in [15, 31] even on experimental platforms given in [32]. One of the main advantage of using model-based controller is fast convergence time which will provide ability to react changes on possibly rough terrains with low error rates.

## 1.2 Spring-loaded Inverted Pendulum Model

An interesting but highly utilized fact about legged locomotor systems is that center of mass trajectories of such behaviours can be described accurately by simple spring–mass models independent of their morphology [33, 34]. Initially designed as a point mass attached to a massless compliant leg in [29], the spring-loaded inverted pendulum (SLIP) model has many variants that are applicable to different legged robot platforms [31, 35–37].

The model is originally motivated by biologic observations given in [38, 39] and various alternatives of SLIP model are mainly used for system identification tools as well as control tools to design input tracking controllers based on the inversion of the Poincaré return maps [4, 15, 28, 40]. The main objective of such controllers is to adapt SLIP model within more complex robotic structures such as the RHex robot given in [41]. Raibert's robots in [1] with the support of similar robots given in [42–44] encourage the idea that SLIP model can be used to regulate running behaviour on robot platforms without the knowledge of its complex structures, see [45, 46].

## 1.3 Approximate Analytical Solutions

Despite the seemingly simple nature of the SLIP model, there are two main problems associated with its equation of motion. First, SLIP has hybrid system dynamics that alternate between flight and stance phases of locomotion. Flight and stance phases can be simply separated from each other by checking whether the foot is on the fly or in contact with the ground, respectively. The remedy for this problem is to derive the equations of motion for each phase separately and switch the phases based on guard functions which detect state-based transition events [29]. The second problem is a more challenging issue for analysing SLIP model. The stance phase of the SLIP model includes non-integrable dynamics preventing the analytic derivation of the equations of motion [47]. An ad-hoc solution for this problem is to use numerical integration to obtain trajectories of the SLIP model numerically. However, numerical integration is a time consuming process and requires huge computational power. As a solution to this problem, some approximations have been proposed to obtain approximate analytical solutions (*A.A.S.*) for the originally non-integrable stance dynamics of the SLIP model [31, 48–51].

Although approximate analytical solutions offer an accurate closed-form representation of the hybrid dynamics of SLIP models, partial feedback linearization is another option that utilize control inputs to eliminate certain non-linear components which appear on the equation of motions given in [52]. [53] uses partial feedback linearization by using another actuator connected to the leg spring in series to eliminate non-linearities combined with hip torque actuation to compensate energy losses during the locomotion. However, physical realization for such solutions usually require complex mechanical design and high energy costs which is not affordable within the revision plan and budget of our one legged robot platform.

Prediction performance of the *A.A.S.*s for the SLIP model has mostly been investigated in simulation studies with a few notable exceptions [4, 32]. However, such experimental validation studies are crucial towards parametric system identification of the robot platforms as well as developing model-based controllers. To this end, practicality of a recent approximate analytic solution to SLIP model with damping has been experimentally validated on a one-legged hopping robot platform [32]. Similarly, practicality of SLIP-based deadbeat controllers have been demonstrated in [4].

## 1.4 Torque-actuation on SLIP Models

One problem about the experimental studies with the legged locomotion models is that there is an inevitable damping loss in the physical systems, which is not originally used in legged locomotion models [48, 54]. Indeed the SLIP model is extended to represent the damping losses in the leg in [31, 55] and its prediction performance on a physical robot platform has been investigated in [32].

One problem with the damping loss is that it is not possible to obtain limit cycle running behaviours with this model if the energy loss is not compensated. To this end, there are some example models that consider hip torque actuation to inject energy to preserve stability of the legged locomotor system, see [15,35,53,56–60]. [35] proposes a clock-driven hip torque actuation for the SLIP model and investigates the stability of the model. Differently, [53] uses hip torque actuation for feedback linearization in order to obtain analytical solutions for the SLIP model. Lastly, [15] proposes an analytical approximate solution for the hip torque actuated SLIP model without increasing the complexity of the approximate stance dynamics solutions of [31].

## 1.5  Key Contributions

The first key contribution of this thesis is the extensions on the TD-SLIP model which provide solutions to the physical problem that is usually neglected on simulation environment. These are necessary for the experimental validation process since the effect of such extensions directly influence COM trajectory on the experimental platform.

Another contribution of this thesis is the revision process and acquired product as a result of it. The revised one legged hopping robot platform is able to represent certain (which can be extended by some mechanical additions) 2D mathematical SLIP models. The robot also provides reliable and accurate physical data thanks to implemented real-time data collection and control system for SLIP models to validate their mathematical properties and to analyse their physical properties that cannot be replicated on simulation environment.

Approximate analytical solution of TD-SLIP model is chosen to be the base of this thesis. Despite promising results of the *A.A.S.* in simulations given in [15], prediction performance of this model in a physical environment is not analysed. One of the key contributions is the experimental validation of the *A.A.S.* of TD-SLIP model. In addition, Ground reaction forces of TD-SLIP model is compared with traditional SLIP model and human running data which provides us interesting results about relation between human data and TD-SLIP model.

As the main aim of this thesis work, a model-based dead-beat controller for the one legged robot platform is designed and implemented by utilizing results of previous contributions. As a result of this implementation, stable and controllable running on the experimental platform is obtained that adapt to the changes on the ground level which is used as a simulation of the rough terrains.

## 1.6   Organization of the Thesis

This thesis work is divided into four main parts which will be explained in detail at the following chapters. In Chapter 2, the main focus will be the investigation of the nature of the SLIP model and analysis of the approximate analytical solution for extended version of the TD-SLIP model in detail. This is a variation of the traditional SLIP model that includes damping losses that are inevitable on physical environments into system dynamics and compensate energy losses caused by damping with the hip torque actuation.

Chapter 3 provides the details about revisions that is done on one legged hoping robot platform on our laboratory. Despite simple adjustments and modifications on the mechanical aspect, electronic structure of the platform is completely changed with a real-time data collection and control system supported by Matlab/Simulink. Design and implementation of this system together with software and hardware solutions about problems that is faced during the process is discussed in detail at Chapter 3.

Prediction performance of the *A.A.S.* will be an essential information for the final part of the thesis. In Chapter 4, we conduct series of systematic experiments on the robot platform to obtain both system parameters and error rate on the *A.A.S.* when working on a physical robot. In addition, ground reaction forces acting on the robot during the stance phase is investigated and compared between human, traditional SLIP model, extended TD-SLIP model and constant torque actuated SLIP model.

Finally, a model-based controller for the robot is designed and implemented on the Chapter 5. Detailed explanation of the design process and implementation of this controller on the robot is given in detail. Chapter 6 concludes our work on this thesis and offers some additional direction for this work on the future.

# Chapter 2

# Approximate Analytical Solution for Extended TD-SLIP Model

In this chapter, we will briefly introduce the well-known spring–loaded inverted pendulum model which will provide base information for analysis and development of the extended TD-SLIP model and its approximate analytical solution. Since the mathematical model will be used for controlling a physical system, some extensions that will increase consistency of the model with robot is done which will be described in detail on following sections. After the implementation of the extensions on the system, we analyse the mathematical derivation of the approximate analytical solution of the extended TD-SLIP model.

## 2.1 Spring–Loaded Inverted Pendulum

Center of mass trajectory of the most legged locomotion systems, independent of their size and morphologies, can be represented by simple spring–mass models. A well known spring–mass system, the Spring-Loaded Inverted Pendulum (SLIP) model originally designed as a point mass attached to a massless compliant leg with no damping during leg compression.

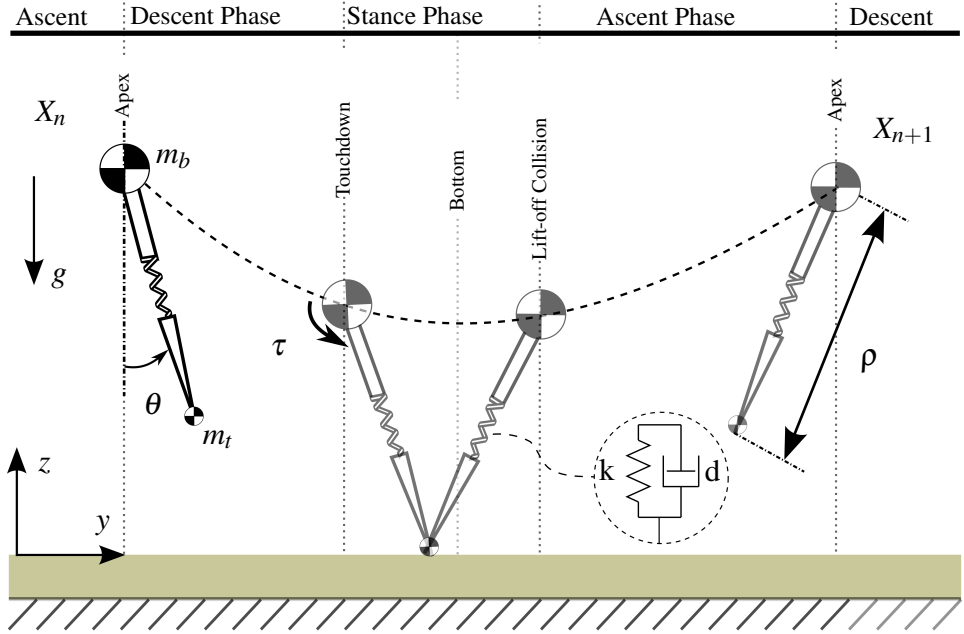

Figure 2.1: Extended TD-SLIP model with detailed illustration of the locomotion phases and corresponding transition events.

However, the effects of damping cannot be neglected on experimental systems, since it will cause an energy loss which leads to inconsistencies on the dynamics of the physical model. As can be seen from Fig. 2.1, when $\tau = 0$ lossy SLIP can be modelled as a point mass, $m$, attached to massless compliant leg with an angle $\theta$ which consist a linear spring with compliance, $k$, viscous damping, $d$.

Due to hybrid system dynamics of the SLIP model that alternates between flight and stance phases, derivation of the equation of motions is done separately by using guard functions to detect certain events on the center of mass trajectory. There are two main events which are touchdown and lift-off events that determine whether system is at stance or flight phases. Touchdown event detects the transition between flight-to-stance which occurs when toe of the leg touches to the ground and lift-off determines the transition between stance-to-flight which occur when the leg loses its contact with ground. During flight phase, apex event is defined as the highest point that model reaches which has a counter part at the stance phase called bottom event where COM trajectory reaches the lowest point. Flight dynamics of the system follows projectile motion which is given as

$$\begin{bmatrix} \ddot{y} \\ \ddot{z} \end{bmatrix} = \begin{bmatrix} 0 \\ -g \end{bmatrix}, \tag{2.1}$$

and the stance dynamics can be obtained by using Lagrangian method , which is given as

$$\frac{d}{dt}\begin{bmatrix} m\dot{\rho} \\ m\rho^2\dot{\theta} \end{bmatrix} = \begin{bmatrix} m\rho\dot{\theta}^2 - mg\cos\theta - k(\rho - \rho_0) - d\dot{\rho} \\ mg\rho\sin\theta \end{bmatrix}. \tag{2.2}$$

As can be seen from (2.2), the SLIP model has non-integrable stance dynamics given in [47] that lead to no exact analytic solutions of the equations of motions. Even tough it is possible to obtain the center of mass trajectories through numerical integration, it requires computing time which can cause problems on real-time experiments. Fortunately, various approximations have been proposed to solve non-integrable stance dynamics of the SLIP model that provide approximate analytical solutions, see [31, 48–51].

## 2.2 Extended TD-SLIP model

The experimental validation of the analytic approximate solution to a torque-actuated dissipative SLIP (TD-SLIP) model will be used to both optimize parameters and check the compatibility of the *A.A.S.* for model-based controller that will be implemented. First, we will investigate the model that will be used throughout the thesis which is an extended version of model given in [15] to be compatible with real-life problems. Fig. 2.1 illustrates the extended TD-SLIP model and system parameters (together with our extensions for physical applicability); body mass ($m_b$), toe mass ($m_t$), spring constant ($k$), damping constant ($d$), leg length ($\rho$), leg rest length ($\rho_0$), vertical and horizontal flight damping ($d_v^f$ and $d_h^f$) and the hip torque ($\tau$). A detailed description of the notation that will be used throughout the thesis can be found in Table 2.1.

Table 2.1: Notation used throughout the thesis

| Extended TD-SLIP Parameters | |
|---|---|
| $y, \dot{y}$ | Horizontal position & velocity |
| $z, \dot{z}$ | Vertical position & velocity |
| $\rho, \rho_0$ | Leg & rest length |
| $\theta$ | Leg angle |
| **Robot parameters** | |
| $m_b$ | Body mass |
| $m_t$ | Toe mass |
| $k, d$ | Linear spring compliance & damping |
| $d_v^f, d_h^f$ | Vertical & horizontal flight damping |

In order to analyse cyclic motions done during the locomotion, a return map should be defined with three different sub-maps as explained in the sequel.

Let $X_n = [z_a^n, \dot{y}_a^n]^T$ denote the apex state at the *nth* stride. By using the flight dynamics given in (2.1), we can find the touchdown state values $\rho_{td}, \theta_{td}, \dot{\theta}_{td}, \dot{\rho}_{td}$. Let us define the descent map $R_d$ as follows:

$$\left[\rho_{td}, \theta_{td}, \dot{\theta}_{td}, \dot{\rho}_{td}\right]^T = R_d \left[z_a^n, \dot{y}_a^n\right]^T. \tag{2.3}$$

By using the values of $\rho_{td}, \theta_{td}, \dot{\theta}_{td}, \dot{\rho}_{td}$ determined from (2.3) and the stance dynamics given in (2.2), we can find the lift-off state values $\rho_{lo}, \theta_{lo}, \dot{\rho}_{lo}, \dot{\theta}_{lo}$. Let us define the stance map $R_s$ as follows:

$$\left[\rho_{lo}, \theta_{lo}, \dot{\rho}_{lo}, \dot{\theta}_{lo}\right]^T = R_s \left[\rho_{td}, \theta_{td}, \dot{\theta}_{td}, \dot{\rho}_{td}\right]^T. \tag{2.4}$$

We note that, although $R_d$ given by (2.3) can be found analytically, the stance map $R_s$ cannot be found analytically due to the non-integrability of the stance dynamics. Finally, using the lift-off state values obtained from (2.4) and the flight dynamics given by (2.1), we can find the next apex state $X_{n+1} = [z_a^{n+1}, \dot{y}_a^{n+1}]^T$. Let us define the ascent map $R_{as}$ as follows:

$$\left[z_a^{n+1}, \dot{y}_a^{n+1}\right]^T = R_s \left[\rho_{lo}, \theta_{lo}, \dot{\rho}_{lo}, \dot{\theta}_{lo}\right]^T. \tag{2.5}$$

By combining (2.3)-(2.5), we can obtain the apex-to-apex return map R as follows

$$X_{n+1} = R(X_n), \tag{2.6}$$

when apex return map $R$ is defined as

$$R = R_{as} \circ R_s \circ R_d. \tag{2.7}$$

The subscripts defined in (2.3)-(2.5), for instance $z_a$, $\rho_{td}$, $\rho_{lo}$, indicates apex, touch-down, lift-off events respectively independent of the parameter used throughout the thesis.

Fig. 2.1 also illustrates a sample single stride behaviour of the extended TD-SLIP model. The cyclic motion of the model can be analysed by observing return maps to given Poincaré section. For the legged locomotion models, we choose this section as the apex state, $X_n$, that corresponds to the highest point in vertical direction during a single stride. Having defined the apex return map for a single stride, the model can be divided into two main phases; flight and stance. The flight phase is when the robot is on the fly and can be divided into two sub-phases as descent and ascent based on decreasing and increasing height, respectively. On the other hand, stance phase refers to duration when the toe of the robot is in contact with the ground. Similarly, the stance phase can also be divided into two sub-phases as compression and decompression based on the decreasing and increasing body velocity. In addition to these, the extended SLIP model includes vertical and horizontal flight damping, lift-off collision map and the toe mass, whose details with mathematical reasoning is explained below in detail.

The flight dynamics for both the descent and ascent maps can be obtained as

$$\begin{bmatrix} \ddot{y} \\ \ddot{z} \end{bmatrix} = \begin{bmatrix} -d_h^f \dot{y} \\ -g - d_v^f \dot{z} \end{bmatrix}. \tag{2.8}$$

Similarly, the Lagrangian dynamics for the stance map can be obtained as

$$\frac{d}{dt} \begin{bmatrix} m\dot{\rho} \\ m\rho^2 \dot{\theta} \end{bmatrix} = \begin{bmatrix} m\rho\dot{\theta}^2 - mg\cos\theta - k(\rho - \rho_0) - d\dot{\rho} \\ mg\rho\sin\theta + \tau \end{bmatrix}. \tag{2.9}$$

Note that we neglect the effect of $d_v^f$ and $d_h^f$ during the stance map, since the body dynamics and the leg damping dominates the small flight damping in this phase. The hip torque $\tau$, that is applied only during the stance phase, has a decreasing ramp profile to ensure *A.A.S.* for the equations of motion as explained in [15]

Finally, the lift-off collision refers to inelastic collision between the robot body and the leg. During the decompression phase, the body accelerates upward and collides with the leg stopper mechanism to lift-off together. We consider this event as an inelastic collision between two different masses and model its effects to system dynamics as an instantaneous change in body velocity using conservation of momentum and kinetic energy properties given as

$$m_b \begin{bmatrix} \dot{y}_b^+ & \dot{z}_b^+ \end{bmatrix}^T + m_t \begin{bmatrix} \dot{y}_t^+ & \dot{z}_t^+ \end{bmatrix}^T := m_b \begin{bmatrix} \dot{y}_b^- & \dot{z}_b^- \end{bmatrix}^T + m_t \begin{bmatrix} \dot{y}_t^- & \dot{z}_t^- \end{bmatrix}^T, \tag{2.10}$$

$$m_b \left( \begin{bmatrix} \dot{y}_b^+ & \dot{z}_b^+ \end{bmatrix}^T \right)^2 + m_t \left( \begin{bmatrix} \dot{y}_t^+ & \dot{z}_t^+ \end{bmatrix}^T \right)^2 := m_b \left( \begin{bmatrix} \dot{y}_b^- & \dot{z}_b^- \end{bmatrix}^T \right)^2 + m_t \left( \begin{bmatrix} \dot{y}_t^- & \dot{z}_t^- \end{bmatrix}^T \right)^2, \tag{2.11}$$

where +,- superscripts indicates pre-collision, post-collision, respectively. By solving these two equations, assuming both toe is at the ground until lift-off event $\begin{bmatrix} \dot{y}_t^+ & \dot{z}_t^+ \end{bmatrix}^T = \begin{bmatrix} 0 & 0 \end{bmatrix}^T$ and final velocities are equal $\begin{bmatrix} \dot{y}_t^- & \dot{z}_t^- \end{bmatrix}^T = \begin{bmatrix} \dot{y}_b^- & \dot{z}_b^- \end{bmatrix}^T$, we obtain instantaneous change on the lift-off velocity as

$$\begin{bmatrix} \dot{y}^+ & \dot{z}^+ \end{bmatrix}^T := \frac{m_b}{m_b + m_t} \begin{bmatrix} \dot{y}^- & \dot{z}^- \end{bmatrix}^T. \qquad (2.12)$$

Note that this is the only place where the small toe mass, $m_t$, is considered in our analysis. A detailed justification about assuming a massless during the locomotion but only the lift-off collision can be found in [32].

### 2.2.1 Equations of Motion for the Flight Phase with Damping

Different than classical SLIP model [29], the extended TD-SLIP model includes vertical and horizontal flight damping, whose dynamics are given in (2.8). The solution for the horizontal position for the flight phase can be obtained as

$$y(t) = \frac{\dot{y}_0}{d_h^f}(1 - e^{-d_h^f t}) + y_0, \qquad (2.13)$$

where $y_0$ and $\dot{y}_0$ represents initial horizontal position and velocity, respectively. Similarly, the solutions for the vertical position is obtained as

$$z(t) = \frac{g}{(d_v^f)^2}(1 - e^{-d_v^f t} - d_v^f t) + \frac{\dot{z}_0}{d_v^f}(1 - e^{-d_v^f t}) + z_0, \qquad (2.14)$$

where $z_0$ and $\dot{z}_0$ corresponds to initial vertical position and velocity, respectively.

Having computed the trajectories for the horizontal and vertical position during the flight phase, the velocities can be simply obtained via analytical derivation of (2.13) and (2.14) as

$$\dot{y}(t) = \dot{y}_0 e^{-d_h^f t}, \qquad (2.15)$$

and

$$\dot{z}(t) = \dot{z}_0 e^{-d_v^f t} - \frac{g}{d_v^f}(1 - e^{-d_v^f t}). \qquad (2.16)$$

After implementation of the extensions on the torque-actuated dissipative SLIP model, we could investigate approximate analytical solution of the extended TD-SLIP model.

### 2.2.2 Approximate Analytical Solution

The stance dynamics of the extended TD-SLIP model, given in (2.9), includes non-integrable terms in its Lagrangian form [47]. Thus, exact analytic solution for the equations of motion is not possible. Motivated by the successful studies on *A.A.S.* to stance dynamics of a variety of SLIP models as in [31, 37, 48, 49], we utilize a recent approximation given in [15] to the solutions of (2.9) towards experimental assessment of the predictive performance. Thus, this section briefly summarizes the approximation method of [15] for the stance dynamics of the hip torque actuated dissipative SLIP model.

Note that when there is no hip torque, which corresponds to $\tau(t) = 0$ in (2.9), a successful approximate analytical solution has been derived in [31] and its experimental validation has been shown in [32]. The key contribution of [15] at this point is that the effect of hip torque can be simply integrated into the approximate analytical solutions of [31] when the hip torque has a previously specified profile such as decreasing ramp during the stance phase.

Table 2.2: Notation used for *A.A.S.*

| **Non-linear Parameters** | |
|---:|:---|
| Angular momentum | $p_\theta := m\rho^2 \dot{\theta}$ |
| The natural frequency | $\hat{\omega}_0 := \sqrt{(\frac{k}{m})^2 + 3(\frac{p_\theta}{(m\rho_0^2)})^2}$ |
| Damping ratio | $\zeta := d/(2m\hat{\omega}_0)$ |
| Damped frequency | $\omega_d := \hat{\omega}_0 \sqrt{1 - \zeta^2}$ |
| The forcing term | $F := -g + \rho_0 \omega_0^2 + 4\rho_0 \omega^2$ |

The approximate analytical solution for the dissipative SLIP model, when $\tau(t) =$

0, relies on two main assumptions; small angular span and small leg compression during the stance phase both of which can be simply satisfied by using a stiff leg spring. Under these assumptions, various quantities are defined at Table 2.2. Then approximate analytical solution of (2.9) is obtained as

$$\rho(t) = Me^{-\zeta\hat{\omega}_0 t}\cos(\omega_d t + \phi_1) + F/\hat{\omega}_0^2, \tag{2.17}$$

$$\dot{\rho}(t) = -M\hat{\omega}_0 e^{-\zeta\hat{\omega}_0 t}\cos(\omega_d t + \phi_1 + \phi_2), \tag{2.18}$$

$$\theta(t) = \theta_{td} + Xt \tag{2.19}$$
$$+ Y(e^{-\zeta\hat{\omega}_0 t}\cos(\omega_d + \phi_1 - \phi_2) - \cos(\phi_1 - \phi_2)),$$

$$\dot{\theta}(t) = 3\omega - 2\omega F/(\rho_0 \hat{\omega}_0^2) \tag{2.20}$$
$$- 2wMe^{-\zeta\hat{\omega}_0 t}\cos(\omega_d t + \phi_1)/(\rho_0),$$

where

$$M := \sqrt{A^2 + B^2}, \tag{2.21}$$
$$\phi_1 := \arctan(-B/A), \tag{2.22}$$
$$\phi_2 := \arctan(-\sqrt{1-\zeta^2}/\zeta), \tag{2.23}$$
$$X := 3\omega - 2\omega F/(\rho_0 \hat{\omega}_0^2), \tag{2.24}$$
$$Y := 2wM/(\rho_0 \hat{\omega}_0), \tag{2.25}$$
$$A := \rho_0 - F/\hat{\omega}_0^2, \tag{2.26}$$
$$B := (\dot{\rho}_t d + \zeta\hat{\omega}_0 A)/\omega_d, \tag{2.27}$$

where the details about the derivations of the approximate analytical solution can be found in [31].

One final step to complete approximate analytical solution is to find an expression for the lift-off time, which will be critical for us, since torque actuation must be vanished before the lift-off event. For the dissipative SLIP model, lift-off occurs when the net force on the body becomes zero during the stance phase which can be expressed as

$$k(\rho_0 - \rho(t_{lo})) - d\dot{\rho}(t_{lo}) = 0. \tag{2.28}$$

Assuming a symmetrical trajectory in the stance i.e. $t_{lo} \approx 2t_b$, an approximate solution for the lift-off time can be found as

$$\begin{aligned} t_{lo} &= (2\pi - \arccos(k(\rho_0 - F/\hat{\omega}_0^2)/(\overline{M}M\exp^{-\zeta\hat{\omega}_0 2t_b})) \\ &- \phi_1 - \phi_3)/\omega_d, \end{aligned} \tag{2.29}$$

where

$$\overline{M} := \sqrt{k^2 - 2kd\hat{\omega}_0 \cos\phi_2 + d^2\hat{\omega}_0^2}, \tag{2.30}$$

$$\phi_3 := \arctan((d\hat{\omega}_0 \sin\phi_2)/(d\hat{\omega}_0 \cos\phi_2 - k)). \tag{2.31}$$

Having completed our derivations for the dissipative SLIP model, we now define the torque profile that will be applied during the stance phase as given in [15]

$$\tau(t) = \begin{cases} \tau_0(1 - \frac{t}{t_f}), & if\ 0 \leq t \leq t_f \\ 0, & if\ t > t_f \end{cases}, \tag{2.32}$$

where $t_f$ represent the time when hip torque will be turned off and $\tau_0$ is the initial value for the decreasing torque profile, which is chosen based on energy that needs to be pumped into the system. There are three major advantages of using decreasing ramp torque profile. First, simple functional dependence on time allows easy incorporation to stance equations. Second, when $t_f$ is chosen as the predicted lift-off time, meaning that the torque will be vanished before the lift-off, premature lift-offs can be simply avoided. Last but not least, decreasing ramp torque profile avoids negative work in the system. Thus, [15] proposes to incorporate the effect of hip torque as a simple correction on angular momentum and utilize the approximate analytical solution of [31] as

$$p_\theta(t) = p_\theta(0) + \int_0^t \tau(\eta)d\eta + \int_0^t mg\rho(\eta)\sin(\theta)(\eta)d\eta. \tag{2.33}$$

Since we investigate the locomotion at discrete steps, apex states, the angular momentum correction equation can be converted to

$$\hat{p}_\theta = p_\theta(0) + \Delta p_\tau + \Delta p_g, \qquad (2.34)$$

where $\Delta p_\tau$ corresponds to the effect of hip torque on the angular momentum. Similarly, $\Delta p_g$ represents another correction on angular momentum to compensate for the effects of non-symmetric steps to the equations of motion as in [61].

Solutions for the hip torque and non-symmetric step corrections can be simply found as

$$\Delta p_\tau = \frac{1}{t_{lo}} \int_0^{t_{lo}} \left( \int_0^{\eta_1} \tau(\eta_2) d\eta_2 \right) d\eta_1 \qquad (2.35)$$

$$= \tau_0 \frac{t_{lo}}{3}, \qquad (2.36)$$

$$\Delta p_g = \frac{mgt_{lo}}{6}(2\rho_o \sin\theta_{td} + r_{lo} \sin\theta_{lo}). \qquad (2.37)$$

By substituting $\hat{p}_\theta$ in all derivations, we obtain a new approximate analytical solution that includes effect of both hip torque and non-symmetric steps.

## 2.3 Conclusion

In this chapter, SLIP model and its variant TD-SLIP model are investigated. Lift-off collusion and flight damping are implemented on mathematical model of the TD-SLIP and approximate analytical solution for the extended TD-SLIP model is obtained which will be tested, optimized and used as a predictor for our model-base controller on the one legged hopping robot platform.

# Chapter 3

# One Legged Hopping Robot Platform

This chapter introduces the one-legged hopping robot platform that we developed at Bilkent University towards experimental validation of our research findings on legged locomotion. The following sections explain the details of the experimental setup, its mechanical design, electronic and software infrastructure including the communication system inside the robot.

## 3.1 Experimental Platform

Note that a one-legged hopping robot platform with a real-time data collection and processing infrastructure can be utilized for many purposes related to legged locomotion, robotics and control theory studies. However, the specific goal of this thesis is to utilize this setup towards experimental validation of an approximate analytical solution to the torque-actued SLIP model. Hence, our introduction of the robot system will be focused around specific properties of such a system. Actually, this thesis does not aim to develop a new setup from scratch but it seeks to develop upon an existing one-legged hopping robot platform in our laboratory, see [14] for details of the previous robot system, to make it applicable for our torque-acuted SLIP model analysis. To this end, the fundamental upgrades that we performed on the robot platform can be summarized as

follows

- We updated the whole electronic infrastructure to support real-time data collection and on-line data processing at 1 $KHz$.

- We implemented a new software infrastrucre using Matlab/Simulink interface both to support real-time analysis using Simulink Real-Time Operating Systems and to facilitate implementation of our existing software on Matlab environment.

- We revised the mechanical system to ensure reliable application of hip torque during the stance phase to make it compatible with the torque-actuated SLIP model.

The rest of this chapter explains the details of these revisions on the one-legged hopping robot platform that we developed in our laboratory.

## 3.2 Mechanical Design

This section introduces the robot platform that we use for our experiments. The robot platform, illustrated in Fig. 3.3, is formed to mimic simple spring-mass systems attached to a non-actuated planarizer with a carbon-fiber boom. Mechanical design of this platform can be divided into three different parts which are: leg part, boom connection and planarizer.

Leg of the robot platform is a simple spring-mass system as can be seen from Fig. 3.1. The main problem with leg mechanism is to let the spring move freely while the leg is fixated to the motor and the rest of the body. Hence, we design and manufacture an aluminium connection part that holds both hip motor shaft and two cylindrical ball-bearings. The ball-bearings are connected to a metal shaft with low surface friction which give leg spring the ability to move free from the rest of the platform. Leg mechanism also has a rubber toe in order to increase friction between leg and ground to prevent robot from slipping. The rest length of the compliant robot leg is measured to be around 20.5 *cm* and two different closed & ground compression spring with compliances 10000 $N/m$ and 4500 $N/m$.

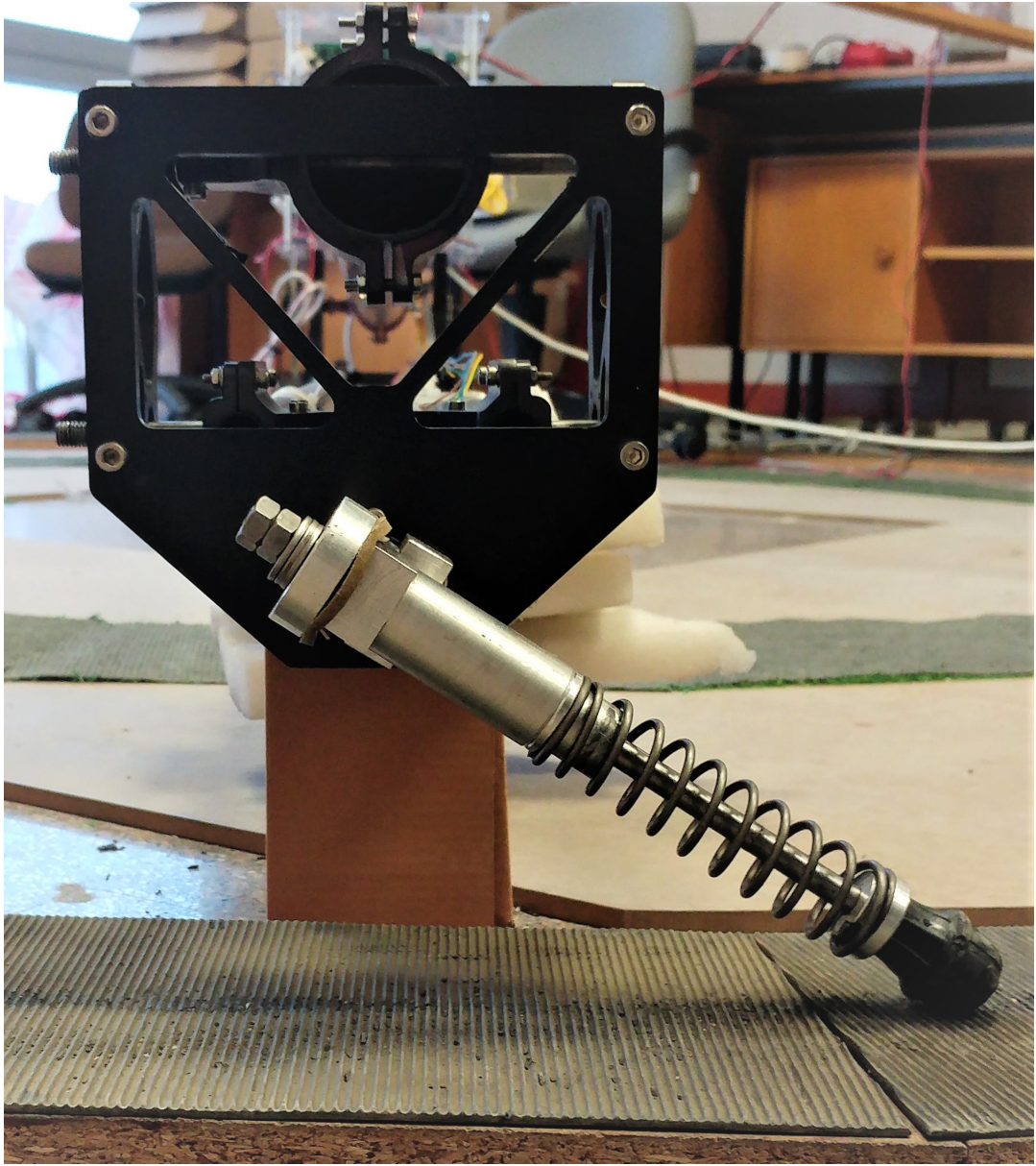

Figure 3.1: Spring mass system attached to the end of the boom.

The boom that connects the robot to the planarizer is 1.67 *m* in length and has $5-cm$ diameter. The robot body is ensured to stay perpendicular to the ground by the use of a $4-bar$ mechanism during the robot locomotion around the planarizer which will fixate the leg to a cylindrical plane with the help of planarizer platform. The robot is equipped with a Maxon EC40-393024 170W brushless DC motor mounted on $4-bar$ mechanism combined with Maxon Planetary Gear-head GP42-C 1:26 and Maxon HEDL 5540 encoder with 500 counts per revolution (CPR) to apply the hip torque during the stance phase and to control the leg angle during the flight phase.

Planarizer platform is used as center for cyclic motion of the leg which uses ball-bearings to maintain low friction while leg traverse in its cylindrical plane. There is no actuators on planarizer but there are two incremental encoders to measure horizontal and vertical position of the robot with 8192 CPR connected to each axis through 1:6 timing belts. Furthermore, all the electronic components are placed onto planarizer to decrease the load on the robot platform.

## 3.3 Electronic Design

The essential part in the revised robot platform is the use of Matlab/Simulink based real time data collection and control architecture. The real-time operating systems (RTOS) guarantees the tasks to be completed in a specified time interval, 1 *Khz* in our case. Matlab offers a soft real-time system which allows tasks to miss pre-specified amount of deadlines. Using a soft real-time system allows us to eliminate total failures caused by ineffective communication delays and exceptions that can occur using multiple hardware devices.

Before describing details of the electronic structure, we need to investigate properties of the operating system that will be used as a basis. Matlab/Simulink real time operating system (SRTOS) consists of two main personal computers (PC) which are:

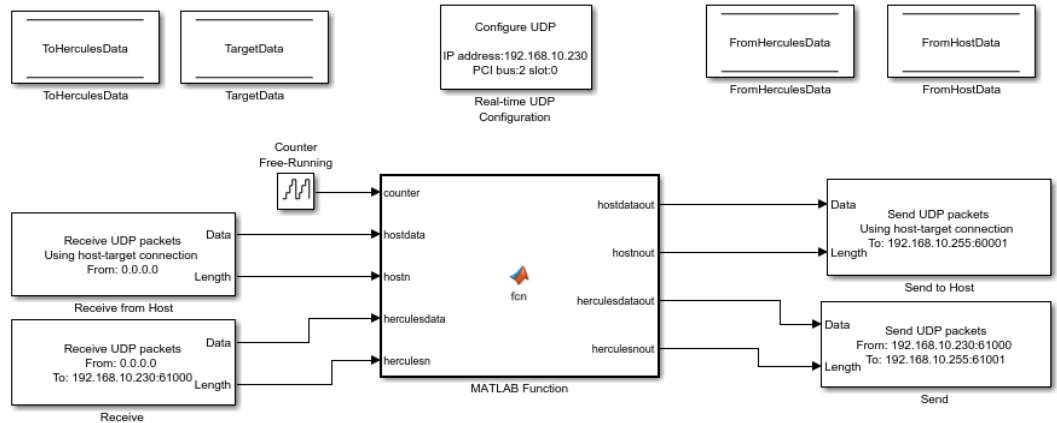

Figure 3.2: Simulink diagram of the one legged hopping robot platform.

Host PC, Target PC. Host PC is used for generating, compiling and embedding the Simulink diagram given in Fig. 3.2 to the target PC. Matlab function block is used as a main block that includes nested functions to gather, process and send the data. In addition, host PC gather the data provided by target PC after processing which includes position, velocity, angle, torque, event, and time information gathered from sensors and hip motor. Matlab/Simulink real-time operating system is implemented on the Target PC and Target PC is used as the main processor for the system such that every information gathered from sensors is processed according to the data and send commands to the hip motors while saving and sending necessary information for the host PC. However, using a personal computer as a main processor can cause compatibility problems that will be discussed further in this chapter.

Electronic structure of this robot platform is based on gathering position data from encoders and controlling motor inputs to manipulate leg angle and hip torque. Although, SRTOS produces high frequency, reliable data for analysis and Matlab environment suggests easy programming interface with wide variety of software options, it only supports a limited amount of hardware for the data acquisition (DAQ). SRTOS offers some DAQ options, however, suggested hardware are mostly not the best cost effective solution in the market which is an essential problem for this platform because of our limited budget. Instead of using single DAQ card connected to target PC, we use a TI Hercules micro-controller which supports different communication protocols and includes built-in encoder reading chips.

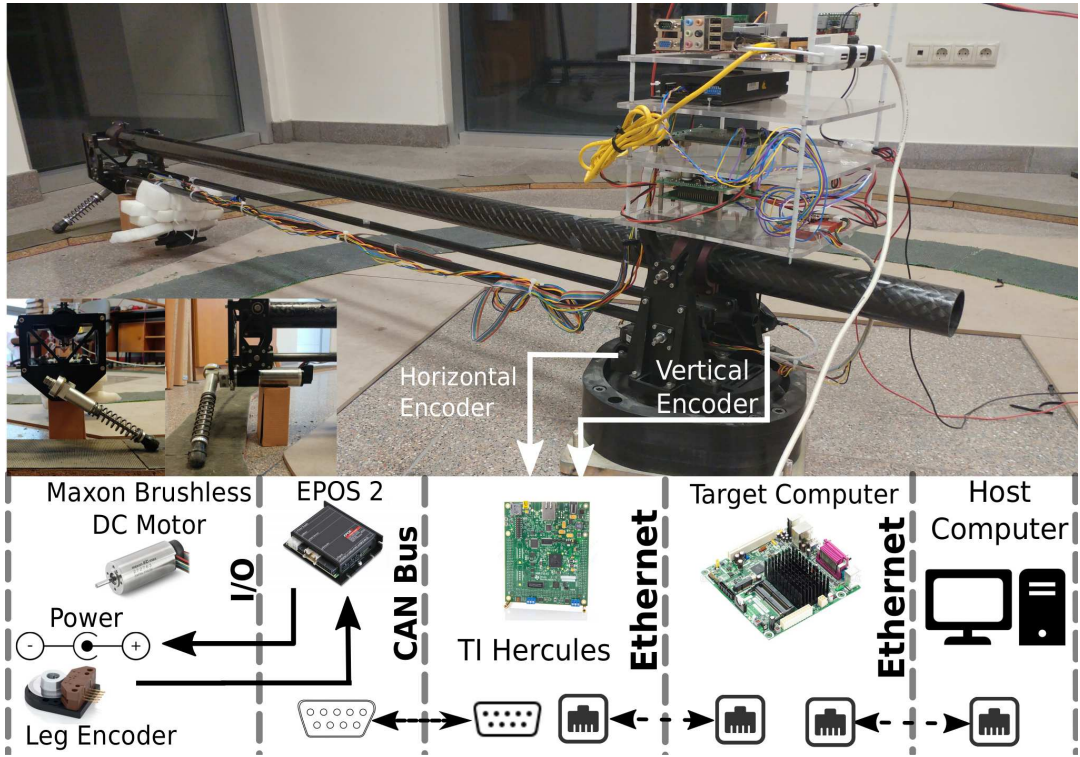

Figure 3.3: The one-legged hopping robot platform used in our experiments together with the electronics and communication infrastructure.

### 3.3.1 Communication Structure

Fig. 3.3 illustrate electronics and communication infrastructure of the robot platform. In order to imitate torque-actuated SLIP model, we need to control both position and current of the hip motor using encoder embedded to Maxon EC40-393024. Consequently, an Epos-2 motor driver is used to both configure motor parameters, precisely read motor encoder and control position, current and velocity of the hip motor.

The main problem with using Epos-2 motor driver is the communication between Epos-2 and the target PC. As we mentioned before, most of CAN PC interface cards are either not supported by MSRTOS or cost-ineffective solutions. Even if a CAN PCI card is available for the robot platform, some additional PCI cards will be required to acquire data from planarizer encoders. To this end, Texas Instruments Hercules microcontroller kit is used as a bridge between hip motor, robot sensors, and target pc to prevent both cost and compatibility problems since TI Hercules supports wide variety of communication protocols.

TI Hercules communicate with hip motor by sending commands given by the target PC using CAN protocol which is faster and more capable than RS-232 serial communication at $1kHz$ fundamental frequency. TI Hercules is programmed to send pre-defined CAN bus commands that is determined by EPOS-2 motor driver, gather encoder information using built-in encoder reader chips and directly send raw data to the target pc for processing using Ethernet communication where PCI cards with more accessible price and variety is available. Hercules is programmed using *C#* programming language which is stand-alone program that does not require modification by MSRTOS.

Raw position data are processed to obtain center of mass coordinates using physical properties of the robot. Position data in Cartesian coordinates is numerically differentiated to obtain velocity data for vertical, horizontal, and leg encoders. Due to noise caused by direct differentiation, a Kalman filter is implemented for vertical and horizontal velocity data. Since vertical, horizontal and leg data are gathered from encoders, the robot is ready for programming applications for parameter identification, ground reaction force, and model-based controller.

## 3.4 Conclusion

To summarize, this section briefly explained the details of the one-legged hopping robot platform that we developed in our laboratory at Bilkent. This robot system will be our test bench for experimental validation of the approximate analytical solution to the torque-actuated SLIP model. The unique real-time data collection and processing system of the current robot platform will enable us to implement different system identification and control algorithm on the robot platform as well for our future research directions.

# Chapter 4

# Parameter Identification and Experimental Validation of the Approximate Analytic Solution for the TD-SLIP Model

The one-legged hopping robot platform is revised on both mechanical and electronics aspects that is described in detail at Chapter 3. In this chapter, We will use the one legged hopping robot to gather data to identify the system parameters of the robot, assess predictive performance for the *A.A.S.* of TD-SLIP model and make a brief analysis and comparison of the ground reaction forces generated by different SLIP models applied on the robot and human data. The main aim of parameter identification is to bring the physical platform to a similar dimension with our mathematical model. By analysing results of this optimization, predictive performance of the *A.A.S.* can be determined that will create a reference point for further applications that will be implemented on the robot and the TD-SLIP model. In addition, analysis on the ground reaction forces created by the robot during stance phase confirms the similarities of the TD-SLIP model with the human running data as well as compares this behaviour with ground reaction forces of lossy SLIP and constant torque-actuated SLIP model.

## 4.1 Data Collection and Pre-process

The system parameters of the robot should be determined before using the *A.A.S.* of the TD-SLIP model in our model-based controller. For this reason, We use a data collection process that is called *"Single-Stride Tests"* to gather apex-to-apex data generated by the robot. This data will be analysed with the data generated by *A.A.S.* that has same initial conditions and control inputs and parameters used in *A.A.S.* are adjusted to obtain minimum apex state errors.

Our experimental validation procedure is similar to [32] which is based on collecting single-stride robot locomotion data with different initial conditions and then assess the predictive performance. To this end, we design a single-stide experiment which consists of five phases which can be seen from Fig. 4.1. *First phase* is the initialization step, where the leg angle is adjusted to the desired touchdown leg angle via a PID controller with a proportional constant, $K_p = 537$, integral constant $K_i = 2179$ and derivative constant, $K_d = 705$ that is determined using auto-tuning option of EPOS-2. Auto-tune feeds motor with pre-defined signals and use the output that is provided by encoders to determine $K_p$, $K_i$, $K_d$ values. We throw the robot upwards for the system to detect a natural apex state in order to avoid any unexpected forces affecting the robot's single stride trajectory. *Second step* is the pre-touchdown phase that corresponds to a very short duration before the touchdown event, which is determined based on robot kinematics to detect touchdown state before 2 *cm* above the ground. At this stage, the PID controller is disabled to avoid any jump backs just before colliding with the ground. We also estimate the lift-off time, $t_{lo}$ given in (2.29) and set the initial torque value, $\tau_0$. *Third step* is the stance phase where the hip motor applies a decreasing ramp torque starting with $\tau_0$ and reaching to 0 at the estimated lift-off instant, $t_{lo}$. *Fourth step* is the ascent phase where the leg angle is fixed with a PID with controller (using the same control parameters with the initialization step) to avoid any distribution on the robot trajectories due to unavoidable leg inertia. Finally, *fifth step* is a brake step after the detection of the second apex state which is blue parts at Fig. 4.1 and data until first apex point and after second apex point (yellow part) is separated from final data at the post-processing. If any failure (red) occur during the test, the corresponding data will be discarded. The robot safely sits back to ground. Note that we only use the data between the two apex states. Therefore, we actually do not use the data from the *fifth step* in our analysis.

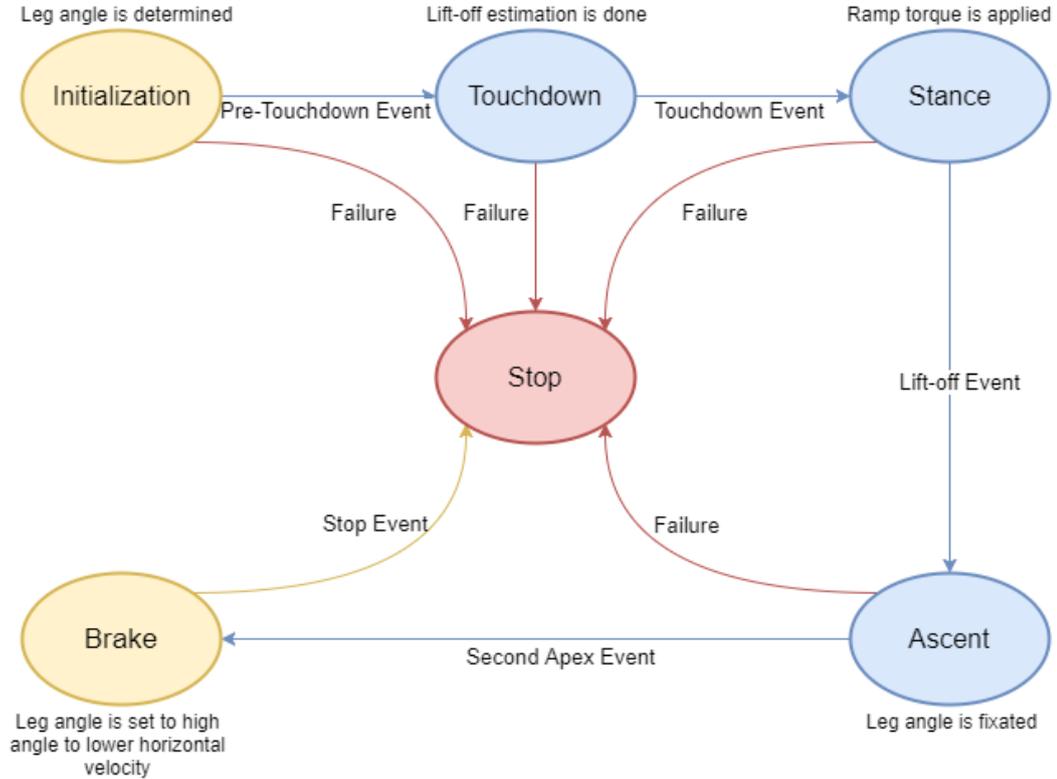

Figure 4.1: Finite state machine diagram of the single stride experiment

During this experiment, we record the 1 *KHz* data coming from DC motor and planarizer encoders for further processing. This data include detailed information about robot state and system such as robot horizontal and vertical position (velocity is obtained through numerical differentiation), leg angle and velocity, motor current and motor torque with respect to 1 *KHz* clock. Before processing this data, we crop apex-to-apex motion and subtract the height of the non-slip ground (2 *cm* height). We then apply an extended Kalman filter (EKF) to reduce the noise in the data both due to resolution of encoders and mostly due to numerical differentiation, see Appendix A.1 for code. The outliers, due to slip of leg during the stance phase etc., are eliminated from the data yielding a total number of successful 120 tests among the 132 experiments.

Fig. 4.2 shows the data gathered from a single stride experiment. As we can see from Fig. 4.2, center of mass trajectory of the robot follows ballistic trajectory until the touchdown event. At the stance phase, vertical velocity decreases until bottom event compressing leg spring to store energy. After bottom event leg spring transfer its energy to the robot which launches from the ground level with the lift-off collusion. At Fig. 4.2.c, we can observe an instantaneous change on vertical velocity. This change is caused by lift-off collusion which occurs between leg structure and robot body, which is added to the mathematical model of our system as an extension as discussed on Chapter 2. Fig. 4.3 illustrates the motor current applied by the hip motor that can be converted to torque $\tau$ by division to the torque constant given as $\tau_c = 396$ and motor gear reduction $G_r = 26$. Because of the direction of the locomotion on the robot graph shows negative current values, however, we use (2.32) which is a decreasing ramp profile starting from touchdown event in this case multiplied by $-1$. The robot uses a PID controller to fixate touchdown angle to pre-defined value until pre-touchdown event as can be seen from Fig. 4.4. Until the lift-off event, leg angle decreases depending on horizontal velocity of the robot and after lift-off event, PID controller enabled to hold the leg at lift-off angle in order to decrease boom oscillations.

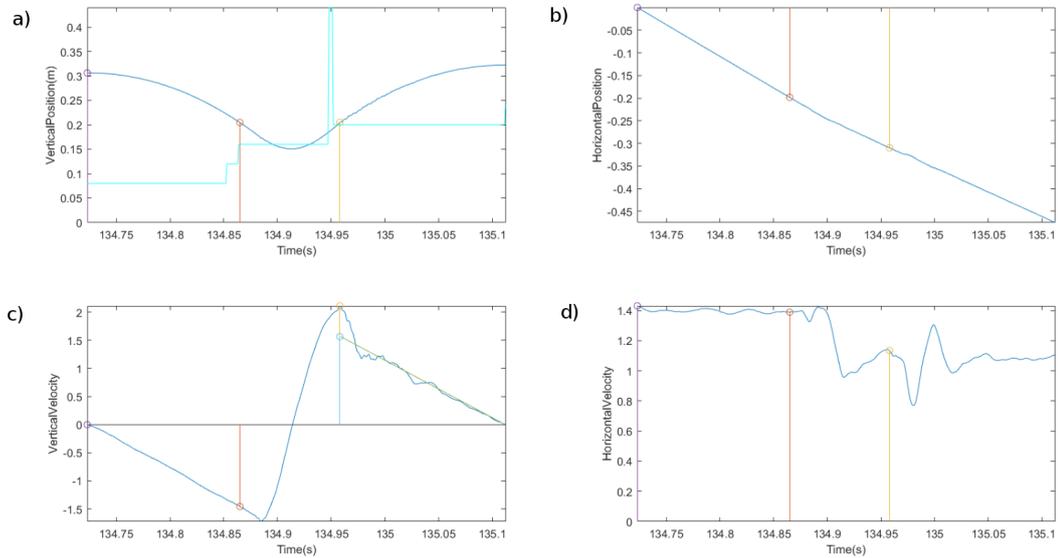

Figure 4.2: Sample single stride position and velocity data taken from robot with phase information *(cyan)*, apex *(purple)*, touchdown*(red)* and lift-off *(yellow)* events: a) Vertical position $z(t)$ b) Horizontal position $y(t)$ c) Vertical velocity $\dot{z}(t)$ and lift-off correction *(green)* d) Horizontal velocity $\dot{y}(t)$

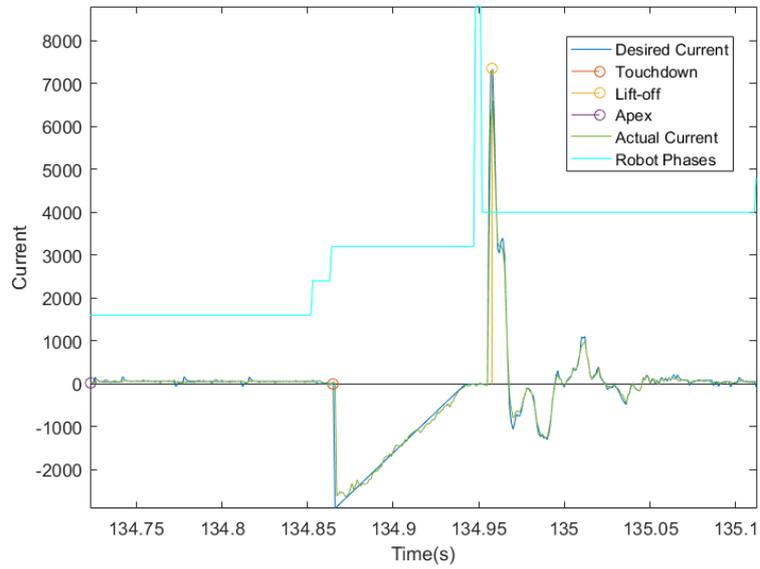

Figure 4.3: Sample single stride motor current data taken from robot with phase information *(cyan)*, apex *(purple)*, touchdown*(red)* and lift-off *(yellow)* events where desired*(blue)* and actual*(magenta)*

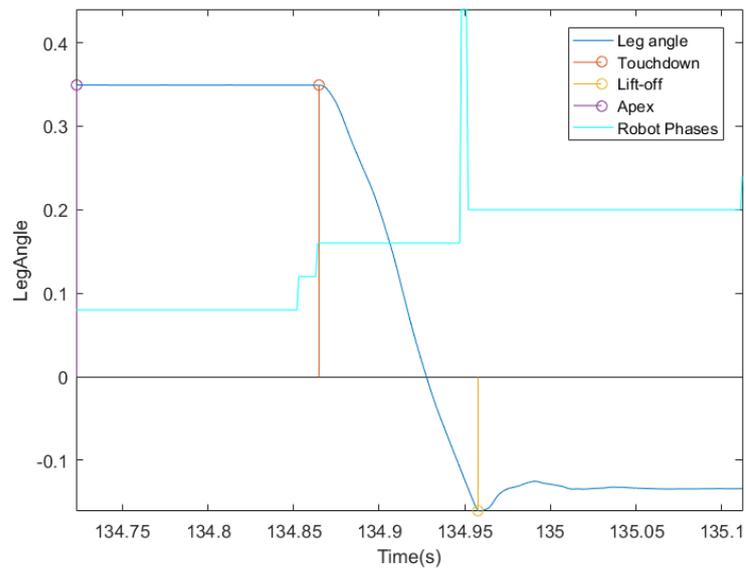

Figure 4.4: Sample single stride leg angle data taken from robot with phase information *(cyan)*, apex *(purple)*, touchdown*(red)* and lift-off *(yellow)* events

## 4.2 Experimental Validation

This section aims to assess the predictive performance of the approximate analytical solution for the TD-SLIP model described in Chapter 2. To accomplish this, we first perform a parametric identification to estimate unknown system parameters that will be used for the approximate analytical solution. To this end, this section first explains our efforts for performing a careful estimation of our robot's system parameters and then assess the predictive performance of the approximate analytical solution via cross-validation to increase statistical significance of our results.

To begin with, we use the 120 successful single stride tests that are collected with our robot as explained in Section 4.1. These experiments are performed with different initial conditions and control parameters. The initial velocity and height values are chosen in the ranges $\dot{y}_0 \in [0.8631, 2.4868]$ $(m/s)$ and $z_0 \in [0.2601, 0.4255]$ $(m)$ to be consistent with [32]. Similarly, the initial torque value and touchdown leg angle are chosen in the range $\tau_0 \in [3,8]$ $Nm$ and to $\theta_{td} \in [10,45]$ $deg$.

As a last step before the identification process, we define three error metrics for apex position, velocity and time error for each stride as

$$E_p = 100 \frac{\left\| \begin{bmatrix} z_a^{n+1} & y_a^{n+1} \end{bmatrix} - \begin{bmatrix} \hat{z}_a^{n+1} & \hat{y}_a^{n+1} \end{bmatrix} \right\|}{\left\| \begin{bmatrix} z_a^{n+1} & y_a^{n+1} \end{bmatrix} \right\|}, \quad (4.1)$$

$$E_v = 100 \frac{\left\| \begin{bmatrix} \dot{y}_a^{n+1} \end{bmatrix} - \begin{bmatrix} \hat{\dot{y}}_a^{n+1} \end{bmatrix} \right\|}{\left\| \begin{bmatrix} \dot{y}_a^{n+1} \end{bmatrix} \right\|}, \quad (4.2)$$

$$E_t = 100 \frac{\left\| \begin{bmatrix} t_a^{n+1} \end{bmatrix} - \begin{bmatrix} \hat{t}_a^{n+1} \end{bmatrix} \right\|}{\left\| \begin{bmatrix} t_a^{n+1} \end{bmatrix} \right\|}, \quad (4.3)$$

where the variables with hats represent the estimated parameters and apex time $t_a$ is determined where $\dot{y}(t) = 0$ during the flight phase. Note that (4.2) does not include vertical velocity, since it is 0 in data by definition of the apex state.

Having defined the error metrics, we utilize an optimization based approach using Nelder-Mead simplex method [62] to perform parametric identification of the robot platform. Our goal is to estimate the parameters $m_b$, $m_t$, $g$, $k$, $d$, $d_v^f$ and $d_h^f$ that minimizes the cost function

$$C = \sqrt{(\frac{1}{N}\sum_{i=1}^{N} E_p^i)^2 + (\frac{1}{N}\sum_{i=1}^{N} E_v^i)^2 + (\frac{1}{N}\sum_{i=1}^{N} E_t^i)^2}, \quad (4.4)$$

where N is the number of different tests. However, instead of using all data in the optimization process, we use 10-fold cross validation to increase statistical significance of our results and to avoid over fitting problems. We separate our data to 10 different sub-sets. One of these is used as a test data (to analyse predictive performance) and remaining ones are used as training data (to estimate system parameters) and the process is repeated for each sub-sets.

## 4.3 Results of Experimental Validation of A.A.S of the TD-SLIP Model

### 4.3.1 System Parameters

System parameters cannot be determined by measurement since they could include effects of different physical properties within them. Compliance of the ground could be included to the spring constant since it will react as a series spring or friction on the robot body can be included to the damping constant. However, we can find approximate values for these parameters and optimize them to obtain satisfactory results for our model-based controller. Table 4.1 presents the estimated system parameters with mean and standard deviations of 10-fold cross validation results. The estimated system parameters are consistent with our robot and expectations based on the results of our previous work, [32].

Table 4.1: System parameters

| **Extended TD-SLIP Estimated System Parameters** | | | | |
|---|---|---|---|---|
| **Parameter** | **Description** | **Mean** | **Std.** | **Units** |
| $m_b$ | Body mass | 2.20 | ±0.06 | $kg$ |
| $m_t$ | Toe mass | 0.03 | ±0.01 | $kg$ |
| $g$ | Corrected gravity† | 11.42 | ±0.07 | $m/s^2$ |
| $k$ | Spring constant | 4696.00 | ±213.50 | $N/m$ |
| $d$ | Damping constant | 9.87 | ±0.60 | $N.s/m$ |
| $d_h^f$ | Horizontal flight damping | 0.01 | ±0.01 | $N.s/m$ |
| $d_v^f$ | Vertical flight damping | 0.23 | ±0.06 | $N.s/m$ |

† See Section 4.3.1.1.

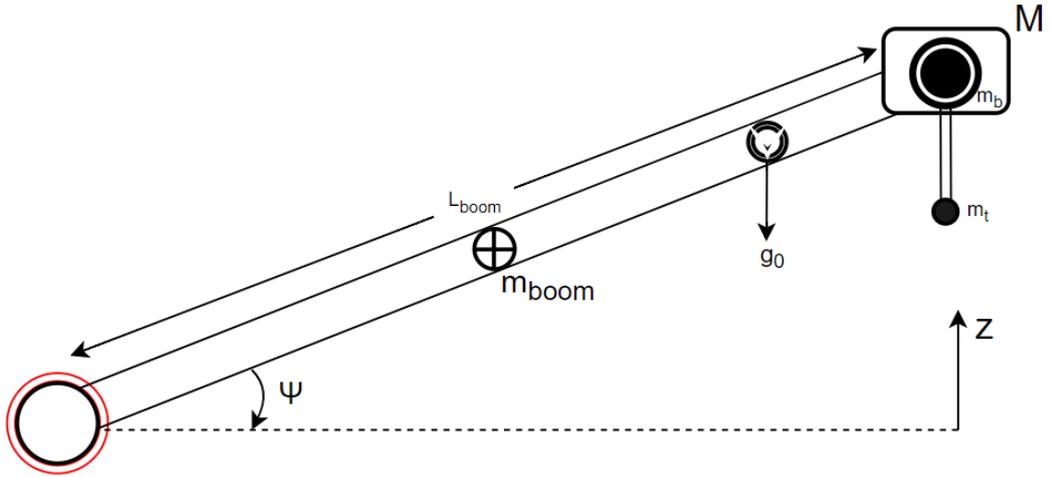

Figure 4.5: Simplified diagram of combined boom and leg structure.

### 4.3.1.1 Boom Dynamics

Note that gravitational acceleration constant, $g$ is also included in the optimization and its estimated value is different than standard value, $g_0 = 9.81 m/s^2$. This is expected since $g = g_0$ is valid for the center of mass of the falling bodies. Since, our robot is attached to the end of a long carbon fiber boom, the robot leg experiences a bigger gravitational acceleration than the center of mass of the boom–robot combination as illustrated in Fig. 4.5. Gravitational acceleration of the leg structure can be calculated with the equations of motions given as

$$(I_{boom} + ML_{boom}^2)\ddot{\psi} = -ML_{boom}g_0\cos\psi - \frac{1}{2}m_{boom}g_0\cos\psi, \quad (4.5)$$

where $M = m_b + m_t = 2.22kg$ is the mass of the leg structure, mass of the boom is defined as $m_{boom} = 0.39kg$, $I_{boom}$ is the inertia of the boom, $\psi$ is the angle of the boom. Assuming boom angle stays small, we can make an approximation $\cos\psi \approx 1$ on (4.5) resulting in

$$(I_{boom} + ML_{boom}^2)\ddot{\psi} = -ML_{boom}g_0 - \frac{1}{2}m_{boom}g_0. \quad (4.6)$$

The conversion from boom angle to vertical robot position can be calculated as $z = L\sin\psi \approx L\psi$ whose second derivative is $\ddot{z} = L\ddot{\psi}$. Combining this conversion with (4.6), we can obtain

$$\ddot{z} \approx \frac{M + m_{boom}/2}{M + m_{boom}/3}g_0, \quad (4.7)$$

where $I_{boom} = m_{boom}L^2/3$ since the tip of the boom is fixated on a cylindrical plane.

Using (4.7), we can obtain a prediction on gravitational acceleration of the leg structure which resulted as $g = 11.41m/s^2$. A similar analysis of this phenomena is given in [32].

### 4.3.2 Predictive Performance

More importantly, Table 4.2 presents the predictive performance of the approximate analytical solution based on cross validation results. We present both the training and test errors of apex position, velocity and time errors for all tests. As seen in Table 4.2, the percentage prediction errors for apex states (position and velocity) is around 5 %, while the percentage prediction error for apex time is around 3 %. Although these results are consistent with [32], predictive performance of the approximate analytical solution for the TD-SLIP is weaker than the one given in [32] for the non-actuated dissipative SLIP model. We believe that the main reason for the bigger prediction error is the integration of hip torque as a simple correction to angular momentum. A more detailed analysis of the effect of hip torque to the system dynamics may reduce the prediction errors for the TD-SLIP model. On the other hand, it has also been shown that such errors can be decreased by the use of model-based adaptive controllers by online identification of the dynamic system parameters [40]. In this case, simple integration of the hip torque to the dissipative SLIP dynamics might be a better option, since simpler models are better for the analysis and control.

Table 4.2: 10-fold Cross Validation Mean Percentage Errors

| Error metrics | Test data | Training data |
|---|---|---|
| $E_p$ | 5.250 ±0.549 | 5.163 ±0.173 |
| $E_v$ | 5.632 ±1.372 | 5.535 ±0.173 |
| $E_t$ | 2.929 ±0.849 | 2.894 ±0.138 |

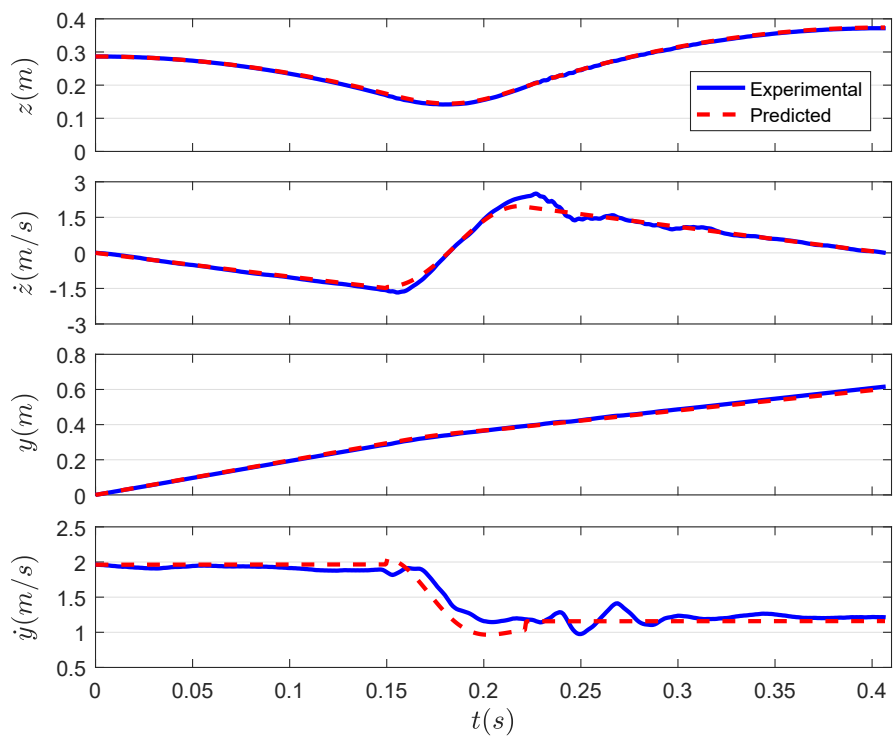

Figure 4.6: Comparison of a sample single-stride experimental data and the approximate analytical solution trajectory.

Last but not least, Fig. 4.6 illustrates a sample single-stride trajectory obtained from the robot platform as well as the predicted trajectory given by the approximate analytical solution. As seen from the figure, the approximate analytical solution does not only predict the next apex states but yields a sufficiently accurate representation for the trajectory of the robot under hip torque actuation. The oscillations in velocity figures on experimental data are due to boom oscillations after the lift-off collision event.

## 4.4 Analysis of Ground Reaction Force

In legged locomotion systems, nature can be used as an example to understand and enhance the capabilities of our models. Recent studies provide a comparison between ground reaction forces (GRF) of running behaviour of humans and predictive SLIP models. Although the analysis of vertical components offers satisfactory results between human running data given at [63] and SLIP model, there are remarkable differences on prediction performance of horizontal components of ground reaction forces. An analysis on torque-actuated dissipative SLIP model illustrates that ground reaction forces of the TD-SLIP model can approximate running behaviour on biological systems sufficiently well as given in [15].

Ground reaction force data are gathered by AMTI Netforce force plate which can provide precise data with 1 kHz frequency as can be seen from Fig. 4.7. Data gathering process is similar to the one described at Section 4.1, however, robot is thrown such that touchdown event occur on force plate which is introduced to the system as a ground offset. Ground level is increased in order to obtain a wide set of initial conditions. Robot is thrown below ground level that allow us to gather low initial apex data. Force data are further processed using simple moving average filters which will prevent the effect of oscillations caused by robot body at force data. However, touchdown and lift-off events cannot be detected exactly by the robot since kinematics of the robot is used to detect such events. For this reason, force data are cropped to obtain better estimation for these events after post-precessing the position and velocity data. For this experiment, we gather 90 tests with different initial conditions $\dot{y}_0 \in [0.95, 3.53]$ $(m/s)$ and $z_0 \in [0.25, 0.41]$ $(m)$ for SLIP, TD-SLIP, TD-SLIP with constant torque while control inputs are varying between $\tau_0 \in [1,9]$ $Nm$ and to $\theta_{td} \in [10,35]$ $deg$.

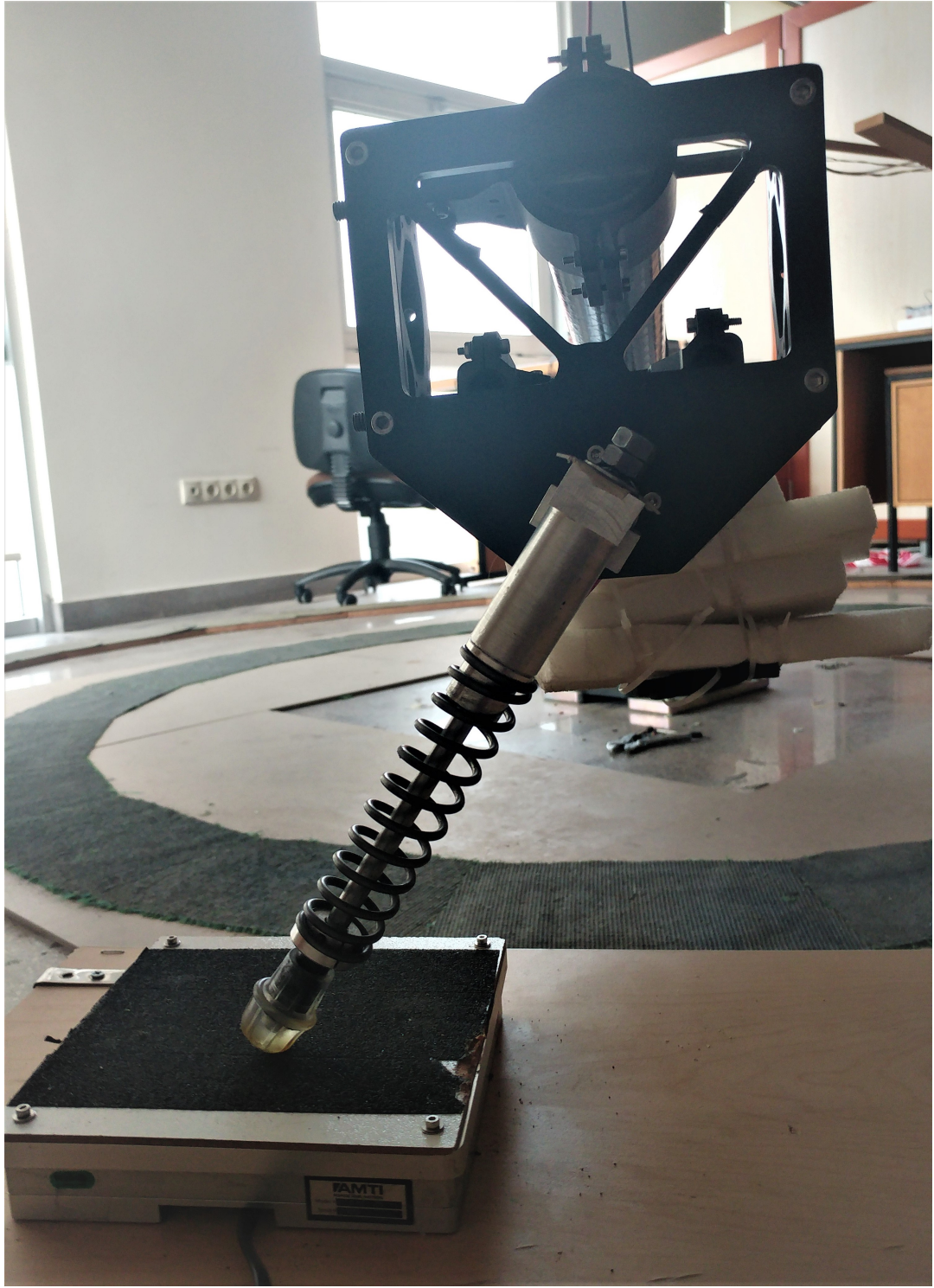

Figure 4.7: Gathering ground reaction force data using AMTI Netforce force plate (Lift-off event)

At Fig. 4.8, center of mass trajectories of stance phase for the sample un-actuated (lossy SLIP), stance-ramp torque (TD-SLIP) and stance-constant torque models analysed with both experimental and theoretical instantaneous ground reaction force vectors called "virtual footfalls" as suggested at [63]. As analysed in [15], simple SLIP model cannot capture backward horizontal forces on the touchdown event, however, using torque actuated models we observe that by supplying energy to the system with both ramp and constant torque profile we can recover backward bias on the horizontal ground reaction forces. Commonly used ideal SLIP model cannot be realized by an experimental platform, however, without any actuation in the system, with the lossy SLIP model, we still cannot recover backward behaviour that occur on the running human data. In this thesis, we try to verify that torque actuated SLIP models may generate GRF data which is sufficiently close to human running GRF data, by considering both theoretical and experimental aspects.

As can be seen at Fig. 4.8, for all given samples force directions generated from robots position data are consistent with experimental data that is gathered from AMTI Netforce force plate for most of the tests. Some differences can be observed close to the touchdown and lift-off event which are mostly caused by parameter adjustments and measurement noises on the hardware. Furthermore, ramp-torque actuated experiments at Fig. 4.8(b) show that TD-SLIP with relatively high initial torque can conveniently replicate ground reaction force directions of human running as given in [63]. Experimental data for ramp torque illustrates that GRF vectors near touchdown event gravitate towards the back of the actual leg location , however, at the end of the stance phase vectors converge to the actual leg location which can also be observed at human running. As different torque profiles utilized on the system, we saw that constant torque profile Fig. 4.9 has tendency to create an additional backward bias at the end of the stance phase which can cause early lift-off conditions on the system as claimed in [15].

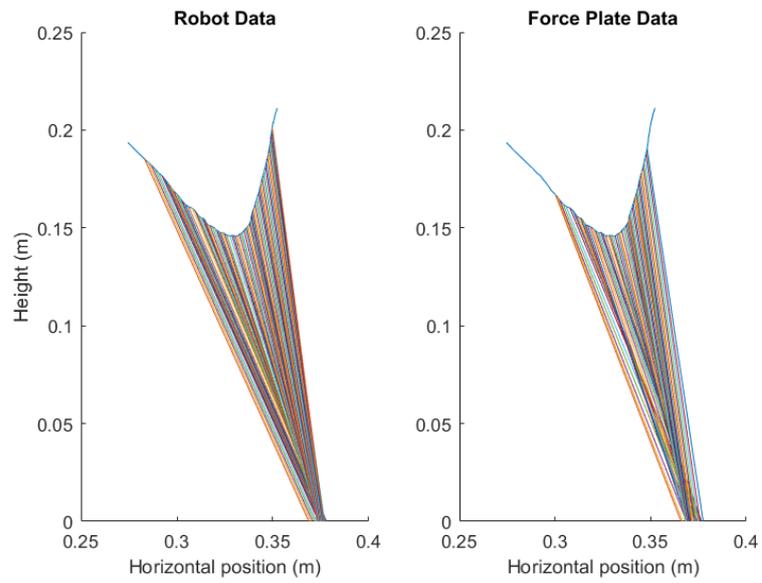

(a) Lossy SLIP model with no actuation

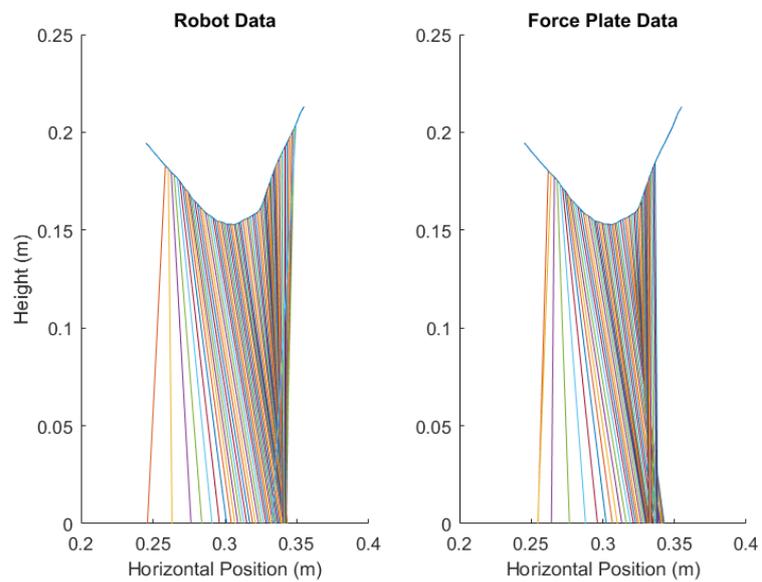

(b) TD-SLIP model with ramp torque actuation

Figure 4.8: Ground reaction force directions for (a) unactuated, (b) ramp torque profiles

Ground reaction force data directly illustrates the effect of hip torque on the robot body which will help us to analyse different torque profiles on the system as well as different leg models. In this thesis work, we examine a brief introduction on ground reaction force analysis. However, in our further research on this topic, we will concentrate on optimizing the hip torque using both natural sources such as human data or with controller using the analysed GRF data.

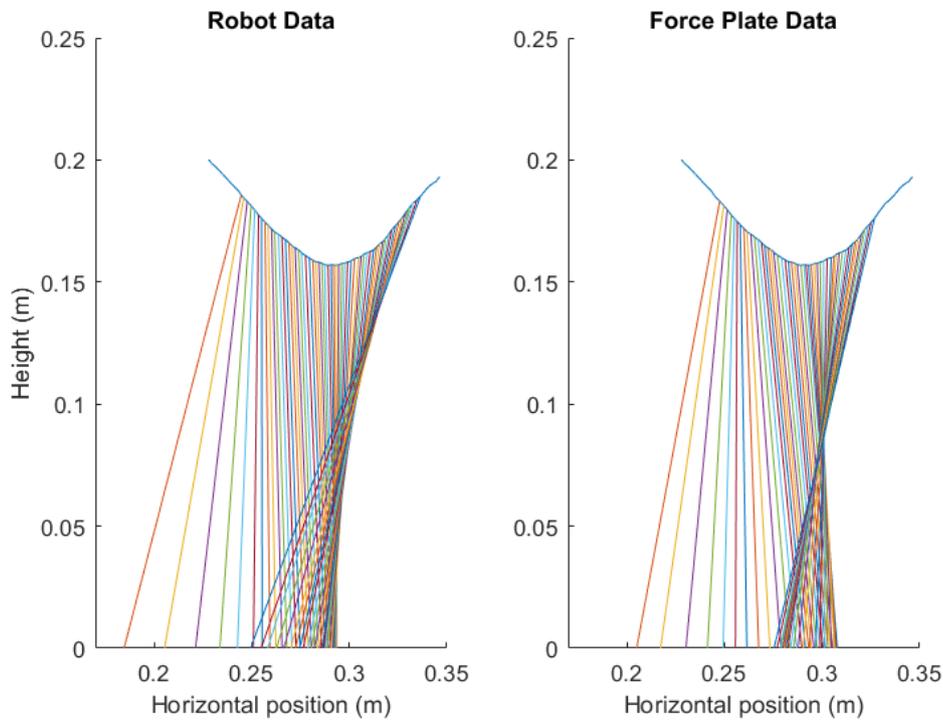

Figure 4.9: Ground reaction force directions for TD-SLIP model with constant torque actuation

## 4.5 Conclusion

In this chapter, we determine a data collection method for the one legged hopping robot. Using data gathered by this method system parameters are optimized by minimizing (4.4) where predicted values are determined by the *A.A.S.* of TD-SLIP model. The parameters obtained with the validation are consistent with the measured ones and with the previous works. As a result of the optimization process, the predictive performance of *A.A.S.* of the TD-SLIP model is assessed. Both next apex position and velocity error are resulted around %5 where next apex time error is determined as approximately %3 which encourage implementation of the model-based controller using *A.A.S.* of the TD-SLIP model on the one legged hopping robot.

# Chapter 5

# Model-Based Controller

One of the main objectives of this thesis is to design a model-based controller and obtain a stable running behaviour on our one legged hoping robot platform. Consequently, TD-SLIP model and its approximate analytical solution are analysed in detail. Secondly, robot platform in our laboratory is revised mechanically to be consistent with the model and electronic structure of the robot is redesigned for accurate and fast operations. Finally, experimental validation of *A.A.S.* is conducted to optimize the system parameters and to assess performance of *A.A.S.* on the experimental platform. Using the results gathered from the previous chapters, we ensure that a model based controller can be designed to regulate vertical height $z_a$ and horizontal velocity $\dot{y}_a$ at the apex event using touchdown leg angle $\theta_{td}$ and initial torque $\tau_0$ as control parameters.

## 5.1 Controller Design

In our controller design, the values of the control inputs $\hat{\tau}_0, \hat{\theta}_{td}$ will be predicted by minimizing error between next apex state values $[z_a^{n+1}, \dot{y}_a^{n+1}]$ and the desired apex state values $[z_a^*, \dot{y}_a^*]$ by using the current apex state values $[z_a^n, \dot{y}_a^n]$. By the inversion of the approximate analytical solution of the TD-SLIP model as apex-to-apex return map, the values of next apex state can be predicted. For this reason, we make an optimization on the *A.A.S.* minimizing the error between $[z_a^{n+1}, \dot{y}_a^{n+1}]$ and $[z_a^*, \dot{y}_a^*]$ by changing $\hat{\tau}_0, \hat{\theta}_{td}$ as we can see in Fig. 5.1. However, computation time of the two dimensional optimization exceeds the system period of $1ms$ and this method doesn't give promising results on the simulation environment because of the correlation between optimization parameters.

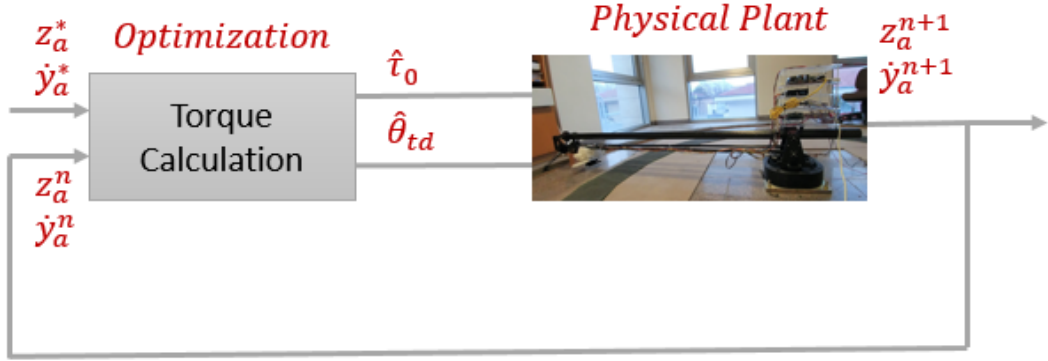

Figure 5.1: Two dimensional model-based controller design diagram

Fig. 5.2 illustrates dead-beat controller design that will be implemented on the experimental platform. Extensively, desired and current vertical position $z_a^*$, $z_a^n$ and horizontal velocity $\dot{y}_a^*$, $\dot{y}_a^n$ are directly used to calculate energy required to reach desired apex states found by

$$E_\tau = \frac{1}{2}m((\dot{z}_a^*)^2 - \dot{z}_a^2) + mg(y_a^* - y_a) + E_{loss}, \tag{5.1}$$

where $E_{loss} := E_d + E_k$ defined as combination of energy loss caused by damping

$$E_d := \int_{t_{lo}}^{0} d\dot{\rho}(\eta)d\eta, \tag{5.2}$$

which can be calculated using *A.A.S.* as

$$E_d := \frac{-d/M^2 \hat{\omega}_0}{4\zeta}(\zeta\cos(2(\phi_1+\phi_2)+\phi_3)+1-$$
$$e^{-2\zeta\hat{\omega}_0 t_{lo}}(\zeta\cos(2\omega_d t_{lo}+2(\phi_1+\phi_2)+\phi_3)+1)), \tag{5.3}$$

and energy left at the leg spring when lift-off event occur before full extension

$$E_k := (\rho_{lo} - \rho_0)^2/2. \tag{5.4}$$

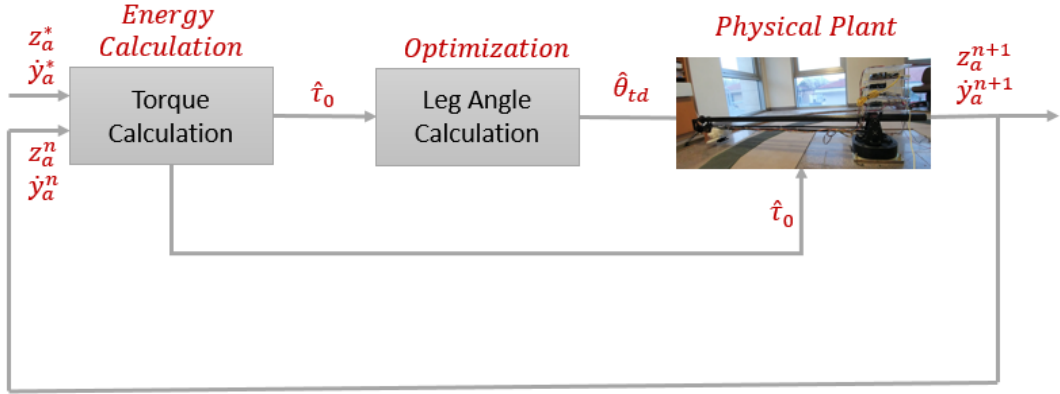

Figure 5.2: Model-based dead-beat controller design diagram of the TD-SLIP model

Energy supplied by hip torque during stance phase is given in [15] as

$$E_\tau := \tau_0 \int_0^{t_{lo}} (1 - \frac{1}{t_{lo}})\dot{\theta}(t)dt, \tag{5.5}$$

where by utilizing approximate return map, (5.1) - (5.5), we can solve for $\tau_0$ and obtain a predicted value $\hat{\tau}_0$ for initial torque value. Using $\hat{\tau}_0$, touchdown leg angle $\hat{\theta}_{td}$ can be predicted by optimizing inverse apex return map according to desired apex states. However, TD-SLIP model does not have an exact analytical return map. Even though, a numerical solution can be used to predict leg angle in simulation environment, robot has a system frequency of 1 $kHz$ which cannot be reached when using a numerical solution. Consequently, *A.A.S.* is used as the apex return map, since, results of the validation in Chapter 4 shows us *A.A.S.* has around %5 error rate for both position and velocity which will not effect the controller drastically.

By inversion of apex return map of TD-SLIP model $\hat{\theta}_{td}$ can be obtained as one dimensional optimisation as follows

$$\hat{\theta}_{td} = \underset{-\frac{\pi}{2}<\theta<\frac{\pi}{2}}{\arg\min}\,(\hat{z}_a^* - (\pi_{\hat{z}_a} \circ R(\theta_{td},[y_a,\hat{z}_a]_k)))^2, \tag{5.6}$$

where we utilize the analytical predictor for return map $R$ defined at (2.7) as numerical solution of the system. After the prediction of the controller parameters, they will be fed to plant to obtain next apex position and velocity $z_a^{n+1}$, $\dot{y}_a^{n+1}$. These values will be used to gather the feedback error in the system which will provide a closed loop control that will conclude our design.

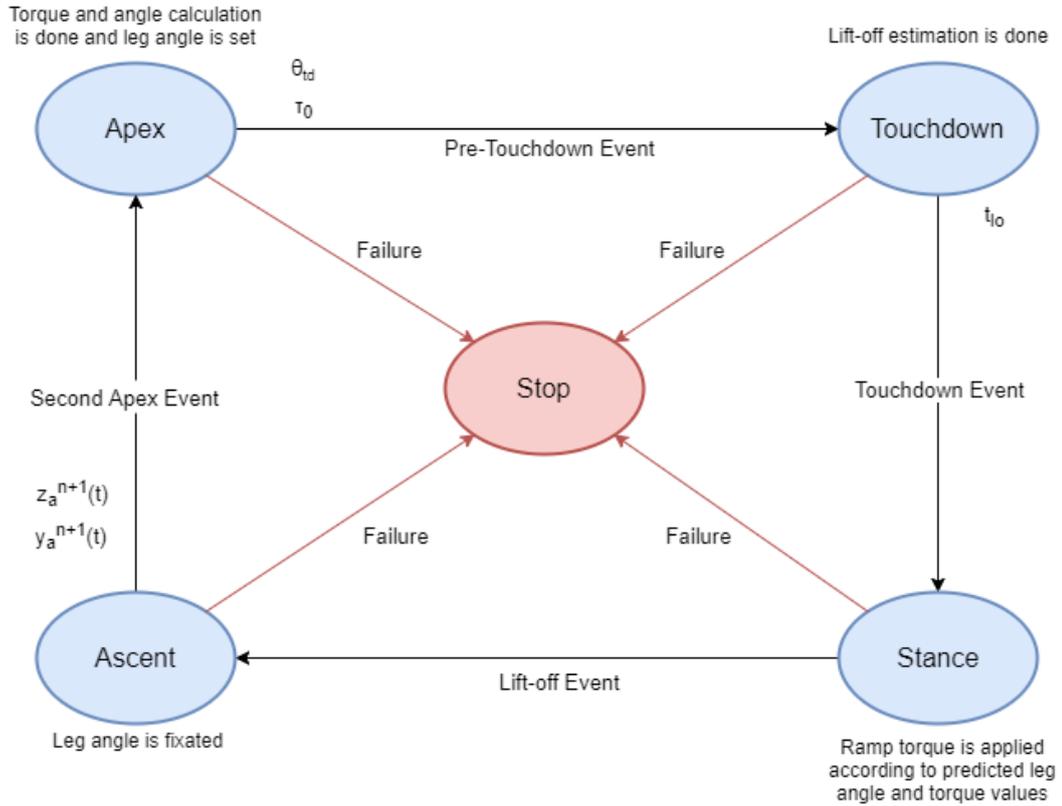

Figure 5.3: Finite state machine diagram of the closed-loop run

## 5.2 Implementation and Results

Controller is implemented on the experimental platform as a finite state machine shown at Fig. 5.3 which uses a similar structure with single stride test Section 4.1. However, state machine returns back to the first phase after reaching next apex event instead of brake phase on the single stride tests. At the first phase after detecting first and next apexes, torque calculation is done since it only requires desired and current apex states. At the second phase, optimisation on the inverse apex return map is implemented using *fminsearch* function provided by Matlab which uses Nelder–Mead simplex search method in low dimensions [62].

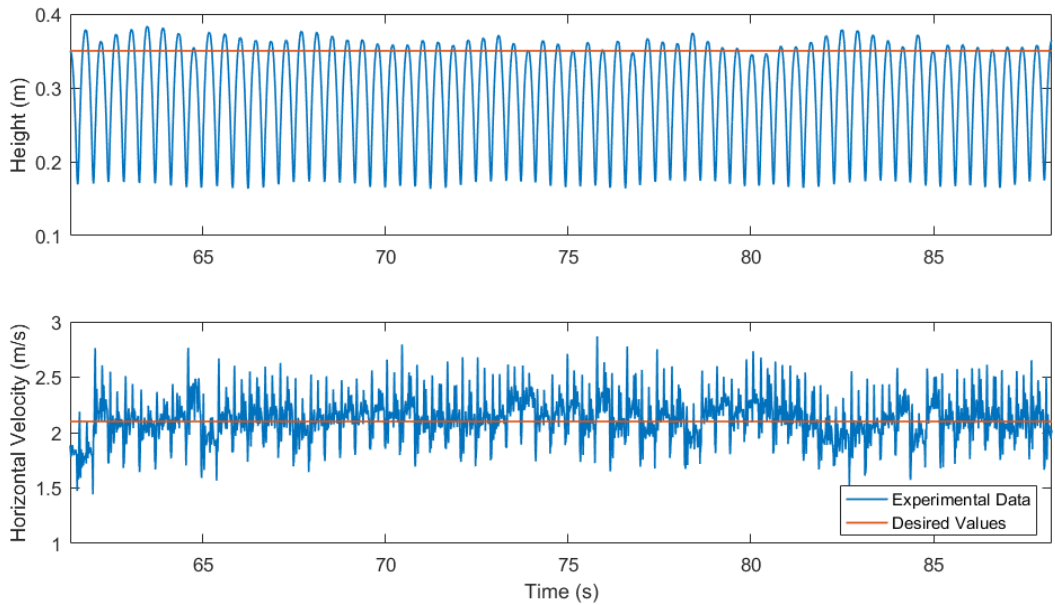

Figure 5.4: Dead-beat controller running data with constant desired values

Fig. 5.4 illustrates an example test run with the implementation of the dead-beat controller on the experimental platform where desired apex height is chosen as $z_a^* = 0.35$ and horizontal velocity is chosen as $\dot{y}_a^* = 2m/s$. As we can observe from this figure, robot can reach both $z_a^*$ and $\dot{y}_a^*$ without failure. Fluctuations at the apex heights are mostly caused by slope of the ground level which can be observed better at the bottom events which cause a sinusoidal disturbance since robot fixated on a cylindrical plane. In addition, we can observe an off-set on both velocity and height data caused by system parameters and prediction performance of the A.A.S. which can be easily solved by adaptive or reactive controllers. As mentioned at previous chapters, the main advantage of the legged locomotion systems is the ability to work on rough terrains which can be introduced to our system as differences on the desired height and velocity. In Fig. 5.5, desired values are changed as

$$t \approx 564 \; s \quad \| \quad z_a^* = 0.39 \; m \rightarrow 0.42 \; m, \tag{5.7}$$

$$t \approx 572 \; s \quad \| \quad z_a^* = 0.42 \; m \rightarrow 0.38 \; m \text{ and } \dot{y}_a^* = 2 \; m/s \rightarrow 2.5 \; m/s. \tag{5.8}$$

and we can see that robot adapts to instantaneous differences on both apex desired height $z_a^*$, horizontal velocity $\dot{y}_a^*$ and quickly converges to the desired values. From these two figures, we can understand that robot is able to perform successive strides whose apex heights can be regulated.

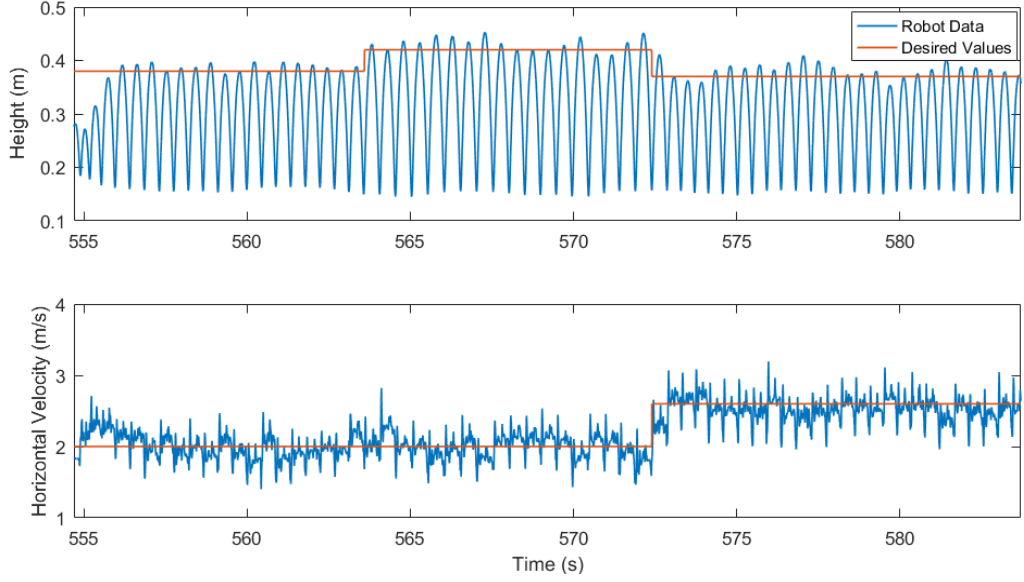

Figure 5.5: Dead-beat controller running data with variable desired values

## 5.3 Conclusion

Inspired by the satisfactory prediction performance of the *A.A.S.* of TD-SLIP model, a model-based controller that determines the control input values by minimizing error between prediction and actual next apex values is proposed. However, two dimensional controller given in Fig. 5.1 exceeds the time constraints determined by our system frequency ($1kHz$). For this reason, design of the controller is converted to a one dimensional optimization given in (5.6) combined with approximation of the initial torque value that is calculated by energy required to reach desired values as can be seen from (5.1) - (5.5). This controller is implemented on our one legged hopping robot platform using a finite state machine given in Fig. 5.3. As a result of this implementation, we obtain a stable and controllable running behaviour on the robot platform given in Fig. 5.4 and rough terrain is also simulated by instantaneous changes on the desired values that the controller quickly converges to modified apex height and velocity.

# Chapter 6

# Conclusion and Future Work

In this thesis, we presented an experimental validation study and model-based controller for a recent approximate analytical solution to torque-actuated dissipative spring-loaded inverted pendulum (TD-SLIP) model given in [15]. To this end, we first presented our revised one-legged hopping robot platform with real-time data collection capability to be consistent with the chosen model. In order to assess the predictive performance, we first performed a parametric identification to carefully estimate physical robot parameters. Then, we applied 10-fold cross-validation to increase statistical significance of our results. The cross validation yielded a prediction error around 5 % for both apex position and velocity.

In addition, we investigated the ground reaction forces acting on one legged hoping robot when different models are used. GRF results showed that lossy SLIP model cannot generate backward movement that can be observed at both human data and TD-SLIP model. For constant torque-actuated SLIP model, an additional backward movement at the lift-off event is observed which can justify that the early lift-off conditions may occur on constant torque case.

Our experimental validation study provided promising results for developing model-based controllers with the approximate analytical solutions for physical robot platforms. Analytic nature of the approximate solutions decreased time complexity in analysis and controller design. Therefore, we implemented a model-based dead-beat controller on our one legged hoping robot using the results of experimental validation process. As a result of this implementation, a stable and controllable running behaviour with the one legged robot platform is obtained. In order to simulate rough terrains, we changed the desired values during the run and we observed that robot could adapt quickly to the new desired values. However, there is a constant error that is caused by both system parameters and the prediction error on approximate analytical solution.

As a future work, first we are planing to develop an adaptive controller on top of the dead-beat controller for online parameter identification while robot is running. Adaptive controllers are claimed to reduce prediction error caused by system parameters as stated in [40]. Even if adaptive controllers show satisfactory results on simulation environment for legged systems, we need the investigate their performance on a physical platform. Secondly, we will further investigate GRF data with different torque actuated SLIP models in order to determine if these types of model can approximate human running better that classical models.

# Appendix A

# Code

## A.1 Kalman Filter Code

### A.1.1 Callback Code

```matlab
function [ ] = CalculateAngles( )
%% Initial paramaters
global TargetData FromHerculesData

persistent kalmanvf kalmanhf

BoomLenght=1.65;
ParalelHeight=0.135;
LegOffSet=0.09;

Δt=1e-3;
sigma_w = 500;
sigma_v = ((1.65*(2*pi/8192)/6));

ΔtHorizontal=1e-3;
sigma_wHorizontal = 100;
sigma_vHorizontal = (((BoomLenght+LegOffSet)*(2*pi/8192)/6));
```

```matlab
18
19  if isempty(TargetData.ElevationOffset)
20      TargetData.ElevationOffset=0;
21  end
22  %% Kalman filter Parameters
23  if isempty(kalmanvf)
24      kalmanvf.Fc = [0 1 0; 0 0 1; 0 0 0];
25      kalmanvf.Gc = zeros(3,1);
26      kalmanvf.Hc = [1, 0, 0];
27      kalmanvf.Qc = [0, 0, 0; 0, 0, 0; 0 0 1]*sigma_w^2;
28      kalmanvf.Rc = sigma_v^2;
29      kalmanvf.Fd = [1, ∆t, ∆t^2/2; 0, 1, ∆t; 0, 0, 1];
30      kalmanvf.Gd = zeros(3,1);
31      kalmanvf.Hd = [1, 0, 0];
32      kalmanvf.Qd = [∆t^5/20, ∆t^4/8, ∆t^3/6; ... ...
33      ∆t^4/8, ∆t^3/3, ∆t^2/2; ...
34      ∆t^3/6, ∆t^2/2, ∆t] * sigma_w^2;
35      kalmanvf.Rd = sigma_v^2/∆t;
36      kalmanvf.xest_post = zeros(3,1);
37      kalmanvf.Pcov_post = zeros(3,3);
38  end
39
40  if isempty(kalmanhf)
41      kalmanhf.Fc = [0 1 0; 0 0 1; 0 0 0];
42      kalmanhf.Gc = zeros(3,1);
43      kalmanhf.Hc = [1, 0, 0];
44      kalmanhf.Qc = [0, 0, 0; 0, 0, 0; 0 0 1]*sigma_wHorizontal^2;
45      kalmanhf.Rc = sigma_vHorizontal^2;
46      kalmanhf.Fd = [1, ∆tHorizontal, ∆tHorizontal^2/2; 0, 1, ∆...
            tHorizontal; 0, 0, 1];
47      kalmanhf.Gd = zeros(3,1);
48      kalmanhf.Hd = [1, 0, 0];
49      kalmanhf.Qd = [∆tHorizontal^5/20, ∆tHorizontal^4/8, ∆...
            tHorizontal^3/6; ... ...
50      ∆tHorizontal^4/8, ∆tHorizontal^3/3, ∆tHorizontal^2/2; ...
51      ∆tHorizontal^3/6, ∆tHorizontal^2/2, ∆tHorizontal] * ...
            sigma_wHorizontal^2;
52      kalmanhf.Rd = sigma_vHorizontal^2/∆tHorizontal;
53      kalmanhf.xest_post = zeros(3,1);
54      kalmanhf.Pcov_post = zeros(3,3);
```

```matlab
55 end
56
57 %% Conversion of Raw Encoder Data to Catesian Coordinates
58 azenc = FromHerculesData.AzimuthEncoder - ...
       TargetData.AzimuthEncoderOffset;
59 azang = azenc*360/6/4/2048;
60 TargetData.AzimuthAngle = azang;
61
62 elenc = FromHerculesData.ElevationEncoder;
63 elang = elenc*360/6/4/2048;
64 elang = elang+TargetData.ElevationOffset; % Correction
65 TargetData.ElevationAngle = mod(elang + 180, 360) - 180;
66
67 TargetData.CenterAngles(2) = (TargetData.ElevationAngle - ...
       TargetData.CenterAngles(1))*1000;
68 TargetData.CenterAngles(1) = TargetData.ElevationAngle;
69
70 TargetData.CenterAngles(4) = (TargetData.AzimuthAngle - ...
       TargetData.CenterAngles(3))*1000;
71 TargetData.CenterAngles(3) = TargetData.AzimuthAngle;
72
73 VerticalPosition = ...
       ParalelHeight+BoomLenght*sin(pi/180*TargetData.ElevationAngle);
74 TargetData.CenterPositions(2) = (VerticalPosition - ...
       TargetData.CenterPositions(1))*1000;
75 TargetData.CenterPositions(1) = VerticalPosition;
76
77 kalmanvf = VerticalKalmanFilter(kalmanvf, VerticalPosition, ...
       0); % Kalman Filter for Vertical speed
78 TargetData.FilteredXVelocity = kalmanvf.xest_post(2);
79
80 HorizontalPosition = (BoomLenght + ...
       LegOffSet)*pi/180*TargetData.AzimuthAngle;
81 TargetData.CenterPositions(4) = ...
       (HorizontalPosition-TargetData.CenterPositions(3))*1000;
82 TargetData.CenterPositions(3) = HorizontalPosition;
83
84 kalmanhf =  HorizontalKalmanFilter(kalmanhf, ...
       HorizontalPosition, 0); % Kalman Filter for Vertical speed
85 TargetData.FilteredYVelocity = kalmanhf.xest_post(2);
```

```matlab
86
87 motorang = FromHerculesData.MotorEncoder - ...
      TargetData.MotorEncoderOffset;
88 motorang = motorang*360*1/26/4/500;
89
90 LegAngle = mod(motorang + 180, 360) - 180;
91 TargetData.LegVelocity = ( mod(LegAngle - TargetData.LegAngle ...
      + 180, 360) - 180 ) * 1000;
92 TargetData.LegAngle = LegAngle;
93
94 end
```

## A.1.2 Function

```matlab
1 function kfstruct = VerticalKalmanFilter(kfstruct, y, u)
2
3 xest_prior = kfstruct.Fd * kfstruct.xest_post + kfstruct.Gd * u;
4
5 Pcov_prior = kfstruct.Fd * kfstruct.Pcov_post * kfstruct.Fd' ...
      + kfstruct.Qd;
6
7 K = Pcov_prior * kfstruct.Hd' / (kfstruct.Hd * Pcov_prior * ...
      kfstruct.Hd' + kfstruct.Rd);
8
9 kfstruct.xest_post = xest_prior + K * (y - kfstruct.Hd * ...
      xest_prior);
10
11 kfstruct.Pcov_post = (eye(length(xest_prior)) - ...
      K*kfstruct.Hd) * Pcov_prior * (eye(length(xest_prior)) - ...
      K*kfstruct.Hd)' + K * kfstruct.Rd * K';
12
13 end
```